%% file: main.tex
\documentclass[11pt]{article}
\usepackage{cite}
\usepackage{amssymb,amsfonts,,amsthm}
\usepackage[ruled,vlined,linesnumbered]{algorithm2e}
\usepackage{graphicx}
\usepackage{textcomp}
\usepackage{subcaption}
\usepackage{fullpage}
\usepackage{style}
\usepackage{enumerate}

\begin{document}
\title{Fast and Robust State Estimation and Tracking \\via Hierarchical Learning}
\author{
  Connor Mclaughlin\thanks{Authors with equal contribution.}\\
    Northeastern University\\
  \texttt{mclaughlin.co@northeastern.edu}\\
  \and
  Matthew Ding\footnotemark[1]\\
    University of California, Berkeley\\
  \texttt{matthewding@berkeley.edu}\\
  \and
  Deniz Erdogmus\\
    Northeastern University\\
  \texttt{d.erdogmus@northeastern.edu}\\
  \and
  Lili Su\\
    Northeastern University\\
  \texttt{l.su@northeastern.edu}
}
\date{September 13, 2024}

\maketitle

\begin{abstract}
Fast and reliable state estimation and tracking are essential for real-time situation awareness in Cyber-Physical Systems (CPS) operating in tactical environments or complicated civilian environments.  
Traditional centralized solutions do not scale well whereas existing fully distributed solutions over large networks suffer slow convergence, and are vulnerable to a wide spectrum of communication failures. 
In this paper, we aim to speed up the convergence and enhance the resilience of state estimation and tracking for large-scale networks using a simple hierarchical system architecture. 

We propose two ``consensus + innovation'' algorithms, both of which rely on a novel hierarchical push-sum consensus component. We characterize their convergence rates under a linear local observation model and minimal technical assumptions. We numerically validate our algorithms through simulation studies of underwater acoustic networks and large-scale synthetic networks. 
\end{abstract}
\newpage 
\tableofcontents
\newpage

\section{Introduction}
\label{sec: intro}
State estimation and tracking are fundamental for the reliable and efficient perception of Cyber-Physical Systems (CPS) operating in tactical environments or complicated civilian environments.    
Many CPS operate in dynamic and uncertain surroundings, exemplified by connected-and-autonomous vehicles (CAVs) \cite{bersani2021integrated,parr2021agent}, industrial internet of things (IIoT) \cite{xu2018survey}, and smart grids \cite{WirelessEstimation, IOTEstimation}. 
By endowing artificial agents within these systems with real-time state estimation and tracking capabilities, they can continually update their understanding of the environment, enabling informed decision-making. 
For example, trajectory planning of autonomous vehicles crucially relies on the accurate and timely perception of the surrounding traffic \cite{peng2023privacy,wang2023towards}.   
In addition, within smart power grids, distributed state estimation uses meters to estimate voltage phasors \cite{SmartGrid1, SmartGrid2}. 
 
The rapid evolution of ubiquitous sensing and communication technologies revolutionizes  CPS with ever-increasing large populations. In modern warfare, large-scale sensors and end/edge devices are deployed across various areas of the battlefield to monitor enemy activities, gather intelligence, and track the positions of friendly forces.  In smart cities, sensors, cameras, and other devices are deployed across the entire city area to collect data on traffic flow, air quality, trash bin status, and streetlight energy consumption. 
Traditional centralized solutions do not scale well.  
Fully distributed solutions have attracted much attention \cite{tsitsiklis1989decentralized,varshney2012distributed,xie2012fully,kar2014distributed}. Yet, as the scale of the multi-agent network increases, existing fully distributed solutions start to lag behind due to crucial real-world challenges such as slow information propagation and network communication failures.
To see the latter, the wearable devices involved in healthcare monitoring systems are not always connected due to low batteries.
Similarly, the vehicles in CAVs are frequently disconnected due to signal blockage or vehicle mobility. 

\begin{figure}
    \centering
    \includegraphics[width=0.4\textwidth]{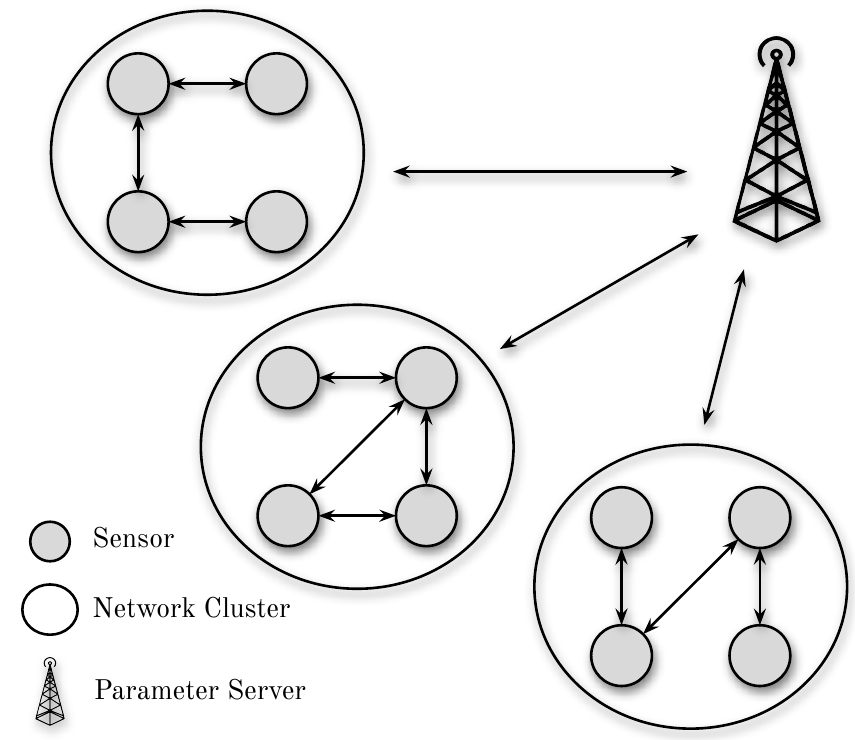}
    \caption{The System Architecture of Hierarchical Learning}
    \label{fig:architecture}
\end{figure}

In this paper, we aim to speed up the convergence and enhance the resilience of state estimation and tracking for large-scale networks using a simple hierarchical system architecture wherein the original networks can be decomposed into small clusters, and a parameter server exists to aid the information exchanges among networks. A simple illustration of the system architecture is given in  Fig.\ref{fig:architecture}. Similar system architectures are considered in~\cite{hierarchical_state, hierarchical_state_2, Hierarchical_FL}. 
We focus on the challenging packet-dropping link failures \cite{7219386} wherein a communication link may drop the transmitted messages unexpectedly and without notifying the sender.  
We do not impose any statistical patterns on the link failures; instead, we only require a link to function properly at least once during a time window.

\vskip 0.5\baselineskip 
\noindent{\bf Contributions.}  
Our contributions are three-fold: 
\begin{itemize}
\item  We propose two hierarchical ``consensus + innovation'' algorithms -- one for state estimation and one for state tracking. 
Both algorithms use a novel hierarchical push-sum consensus component (i.e., Algorithm \ref{alg:push-sum hierarchical FL}) that enables communication-efficient synchronization (one agent per sub-network) across sub-networks with a tuning parameter that controls the synchronization frequency.   
\item We characterize the convergence rates of both of the algorithms under a linear local observation model. As can be seen from our analysis, the worst-case convergence speed depends on the diameters of the subnetworks only, which are significantly smaller provided the original network can be well partitioned into smaller subnetworks.  
In addition, with hierarchical learning, the original networks are no longer required to be strongly connected.  
\item Finally, we provide numerical results to validate our theory. 
We empirically validate the robustness and improved convergence of our algorithms through simulations of large-scale networks and an application of underwater acoustic networks, where our method results in over 30\% faster convergence in realistic system configurations.  
\end{itemize}

\section{Related Work} 

\subsection{Hierarchical algorithms}
Hierarchical algorithms have been considered in literature~\cite{hierarchical_state, hierarchical_state_2,EPSTEIN2008612,hou2019groupinformation}.   
Focusing on electric power systems, a manager-worker architecture was considered in early works ~\cite{hierarchical_state, hierarchical_state_2},  which decomposed a large-scale composite system into subsystems that work in the orchestra with a central server. 
Carefully examining the dynamics in electric power systems, both~\cite{hierarchical_state} and~\cite{hierarchical_state_2} proposed two-level hierarchical algorithms with well-calibrated local pre-processing in the first level and effective one-shot aggregation in the second level. 
Different from \cite{hierarchical_state, hierarchical_state_2}, we go beyond electric power systems and our algorithms do not require complicated pre-processing. 

Compared with a single large network, using hierarchical system architecture to speed up convergence was considered~\cite{EPSTEIN2008612,hou2019groupinformation}.  
Epstein et al.\cite{EPSTEIN2008612} studied the simpler problem of average consensus, and mathematically analyzed the benefits in convergence speed. 
However, there is a non-diminishing term in their error convergence rate (see \cite[Theorem 8]{EPSTEIN2008612} for details). As can be seen from our analysis, for consensus-based distributed optimization algorithms to work, it is important to guarantee consensus errors quickly decay to zero because as the algorithm executes the consensus errors for the (stochastic) gradients in each round will be accumulated. Hou and Zheng~\cite{hou2019groupinformation} used a hierarchical ``clustered'' view for average consensus. Nodes are clustered into small groups, and at each iteration receive estimates from peers within the same group as well as group information of other groups. Here, the group information is defined as a weighted combination of the local estimates of the group members. Unfortunately, such information is often expensive to obtain. Fujimori et al.~\cite{fujimori2011sice} considered achieving consensus with local nonlinear controls wherein the notion of consensus considered departs from the classical average consensus. Specifically, with their methods, the value that each node agrees on may not be the average of the initial conditions/values; see the numerical results in \cite{fujimori2011sice} for an instance in this regard. 
Wang et al.~\cite{wang2016consensustracking} considered the consensus tracking problem and designed a well-calibrated fusion strategy at the central server. Yet, this method requires that on average a non-trivial portion of the states are observable locally, which does not hold in our setup.

\subsection{Fault-tolerant algorithms}
Enabling fault tolerance is fundamental for computing with large-scale distributed systems \cite{Lynch:1996:DA:2821576,tsitsiklis1986distributed,lamportbyzantine,Lyapunov}.   

Hadjicostis et al.~\cite{7219386} studied the practical yet challenging packet-dropping link failures where the packet drops across different links or times steps are not necessarily independent.
Building on top of the well-known push-sum method \cite{kempe2003gossip}, they proposed an algorithm under which each agent exchanges a set of running sums with its immediate neighbors and showed the convergence by establishing an equivalence by running a standard push-sum on an augmented graph whose links are fault-free.  
This approach has inspired numerous subsequent works, including resilience to Byzantine agents through a robust aggregation step ~\cite{SU2017352}, and tighter, network-independent bounds for strongly convex problems~\cite{spiridonoff2020robust}. 

In the control literature, Lyapunov-based methods~\cite{Lyapunov} have been powerful tools in ensuring system stability for a range of in-system imperfections such as time-delays~\cite{liu2023adaptive} and dynamics uncertainties~\cite{wang23tac}.  
For example, Liu et al.~\cite{liu2023adaptive} used a Lyapunov-Krasovskii functional approach to handle unknown communication delays.
To handle dynamics uncertainties~\cite{wang23tac}, they proposed an algorithm that recursively applies a Lyapunov function, virtual control laws, and tuning functions at each step to adaptively update the controller. 

Similar to the first line of work, we consider packet-dropping link failures. Differentiating from both lines of these works, we study the inference and tracking problem. 

\subsection{Comparison with Previous Work} 
On the technical side, \cite{spiridonoff2020robust} and \cite{UnreliableConvergence} are closest to our work. 

Spiridonoff et al.\,\cite{spiridonoff2020robust} considered the general distributed optimization in the presence of packet-dropping links, agent activation asynchrony, and network communication delay. They used similar algorithmic techniques as in~\cite{vaidya2012robust} to achieve resilience. Different from \cite{spiridonoff2020robust}, we consider the concrete state estimation and tracking problem and less harsh network environment (i.e., synchronous updates and no communication delay). Nevertheless, we manage to relax the strong-convexity assumption in \cite{spiridonoff2020robust} and consider time-varying global objectives. To the best of our knowledge, algorithm resilience in hierarchical systems against non-benign network failures is largely overlooked.

Departing from \cite{UnreliableConvergence}, we study the state estimation and tracking problems. Our work overlaps with \cite{UnreliableConvergence} for the special case where the system contains only one network or the synchronization frequency among sub-networks is 1. In addition, \cite{UnreliableConvergence} focuses on the pure optimization problem assuming exact knowledge of local gradients.
In contrast, we study an inference problem (i.e., recovering the underlying truth in the presence of noises). 
In terms of analysis techniques, though our work shares overlap with \cite{UnreliableConvergence} in the use of the underlying consensus primitive, our analysis on the general collaborative tracking is significantly different from reference \cite{UnreliableConvergence}. 
Specifically, we need to ensure the update of local cumulative quantities complies with the global state dynamics. The resulting analysis involves dealing with sequences of Kronecker matrix products, eigenvalue decomposition, and matrix concentration. 

\section{Problem Formulation}
\subsection{System Model}
\label{subsec: system model}
Let $G(\calV, \calE[t])$ be a multi-agent network with  $|\calV| =N$, directed edges, and time-varying edge set $\calE[t]$.  
Let $\{G(\calV_i, \calE_i[t])\}_{i=1}^M$ be any given decomposition of $G(\calV, \calE[t])$ such that $|\calV_i| = n_i$, $\sum_{i=1}^M n_i = N$, and $\cup_{i=1}^M \calE_i[t] \subseteq \calE[t]$.  There are many graph decomposition algorithms such as spectral clustering algorithms \cite{rohe2011spectral,chen2014improved,andersen2006local} and algorithms that find strongly connected components \cite{cormen2022introduction}.  However, the specific selection or construction of such an algorithm is beyond the scope of this paper.
We assume there exists a parameter server (PS) that aids the information exchanges among sub-networks $G(\calV_i, \calE_i[t])$. Similar system architecture is adopted in the literature~\cite{edge1, edge2,Hierarchical_FL}. Though the edge set of a sub-network $G(\calV_i, \calE_i[t])$ can be time-varying, we  assume that there exists $\calE_i$ such that $\calE_i[t]\subseteq \calE_i$ for each $t$.

Let $j\in \calV_i$. 
Denote $\calI_j^i[t] = \{k \mid (k,j) \in \calE_i[t] \}$  
and $\calO_j^i[t] = \{k \mid (j,k) \in \calE_i[t] \}$ as the sets of incoming and outgoing neighbors to agent $j$, respectively. 
For notation convenience, we denote $d_j^i[t] = \abth{O_j^i[t]}$.\footnote{This will not create confusion because only $\abth{O_j^i[t]}$ will be used in the algorithm.}

For ease of exposition, we assume agents in the same sub-network can exchange messages subject to the local network structure $G(\calV_i, \calE_i[t])$. No messages can be exchanged directly between agents in different sub-networks. In addition,  the PS has the freedom to query and push messages to any agent. Nevertheless, such message exchange is costly and needs to be sparse. We conjecture that, with careful control of weight splitting to ensure mass preservation across sub-networks, our results can be extended to the more general settings wherein agents across sub-networks can exchange messages when connected. We leave this to future work.  

\vskip 0.5\baselineskip 

Throughout this paper, we use the terminology ``node'' and ``agent'' interchangeably. 
 
\subsection{Threat Model}
\label{subsec: threat model} 
We follow the network fault model adopted in \cite{UnreliableConvergence} to consider packet-dropping link failures. 
Specifically, any communication link may unexpectedly drop a packet transmitted through it, and the sender is unaware of such packet lost. 
If a link successfully delivers messages at communication round $t$, we say this link is {\em operational} at round $t$.
\begin{assumption} 
\label{ass: link reliability}
We assume that a link in $\calE_i$ is operational at least once every $B$ communication round, for some positive constant $B$ for each $i=1, \cdots, M$. 
\end{assumption}

A similar assumption is adopted in \cite{RobustAverageConsensus,spiridonoff2020robust,UnreliableConvergence}. In a sense, imposing a bound on the failure lasting time is necessary. To see this, consider the extreme scenario where all the links fail permanently. Clearly, no information exchanges in the networks. Hence, no learning can be achieved.

\vskip 0.3\baselineskip 
\begin{remark} 
As observed in \cite{RobustAverageConsensus,spiridonoff2020robust,UnreliableConvergence}, the above threat model is much harder to tackle compared with the ones wherein each agent is aware of the message delivery status.  
Yet, this threat model is practically relevant for many applications. 
In harsh and versatile deployment environments such as undersea, 
the communication channels between two neighboring entities may suffer strong interference, leading to unsuccessful message delivery.  
\end{remark}

\subsection{State Dynamics and Local Observation Models} 
\label{subsec: state dynamics + local observation} 
\paragraph{State dynamics}
Each of the agents is interested in learning the $d$-dimensional state of a moving target $w^*[t]$ that follows the dynamics 
\begin{equation}  \label{eq: truth dynamics}
w^*[t] = A w^*[t-1] ~~~ \forall ~ t\ge 1,  
\end{equation}
where 
$A\in \reals^{d\times d}$ is the dynamic matrix that
is known to each agent, and $w^*[t]\in \calW\subseteq \reals^d$. 
For example, each autonomous vehicle needs to keep track of neighboring vehicles, where the global state contains the statuses of the spatial position, the velocity, and the acceleration of the target. In this case, the state dynamic is approximately linear.  

An interesting special case of Eq.\eqref{eq: truth dynamics} is  when $A=I$, under which the target state is time-invariant, i.e., $w^*[t] = w^*[0]$ for all $t$. This special case is often referred to as the state estimation problem  \cite{SmartGrid1}.

\paragraph{Local observation}
In every iteration $t$, each agent locally takes measurements of the underlying truth $w^*[t]$. We focus on the linear observation model which is commonly adopted in literature \cite{FiniteTimeGuarantee, NonlinearObservation, ConsensusStochasticApprox}:
For a specific agent $j\in \calV_i$ at time $t$:
 \begin{equation} \label{eqn: local observation}
    y_j^i[t] := H^i_jw^*[t] + \xi_j^i[t]
 \end{equation} where $H_j^i$ is the local observation matrix, and  $\xi^i_j[t]$ is the observation noise.  
We assume that the observation noise $\xi^i_j[t]$ is independent across time $t$ and across agents $j$. In addition, $\expect{(\xi^i_j[t])^{\top}\xi^i_j[t]}\le \sigma^i_j$.
In practice, the observation matrix $H^i_j$ is often fat. Thus, to correctly estimate/track $w^*[t]$, agents must collaborate with others.

\section{Hierarchical Average Consensus in the Presence of Packet-dropping Failures}
\label{sec: average consensus}
We present an average consensus algorithm (Algorithm \ref{alg:push-sum hierarchical FL}) that builds upon our previous work \cite{UnreliableConvergence}, specifically tailored to the hierarchical system architecture, aiming for guaranteed and rapid finite-time convergence.  

Up to line 13 in Algorithm \ref{alg:push-sum hierarchical FL} is the parallel execution of the fast robust push-sum \cite{UnreliableConvergence} over the $M$ subnetworks. 
Lines 14-23 describe the novel information fusion across the subnetworks, which only occurs once every $\Gamma$ iterations.
\begin{algorithm}
\caption{Hierarchical Push-Sum (HPS)}
\label{alg:push-sum hierarchical FL}
 {\em Initialization}: 
For each sub-network $i=1, \cdots, M$: $z_j^i[0]=w_j^i\in \reals^d$, $m^i_j[0]=1\in \reals,$ 
$\sigma_j^i[0]={\bf 0}\in \reals^d$, $\tilde{\sigma}_j^i[0]=0\in \reals$, and $\rho^i_{j^{\prime}j}[0]={\bf 0}\in \reals^d$, $\tilde{\rho}^i_{j^{\prime}j}[0]=0\in \reals$ for each incoming link, i.e., $j^{\prime} \in \calI_j^i$.

\vskip 0.2\baselineskip 
In parallel, each agent in parallel does:\\
\For{$t\ge 1$}
{ 
$\sigma_j^{i+}[t]  \gets  \sigma_j^i[t-1] + \frac{z_j^i[t-1]}{d_j^i[t]+1}$,
$\tilde{\sigma}_j^{i+}[t] \gets \tilde{\sigma}_j^i[t-1] + \frac{m_j^i[t-1]}{d_j^i[t]+1}$\;

Broadcast $\pth{\sigma^{i+}_j[t], \tilde{\sigma}^{i+}_j[t]}$ to outgoing neighbors\;

\For {each incoming link $(j^{\prime},j)\in \calE_i[t]$}
{\eIf{message $\pth{\sigma^{i+}_{j^{\prime}}[t], \tilde{\sigma}^{i+}_{j^{\prime}}[t]}$ is received}
{$\rho^i_{j^{\prime}j}[t] \gets \sigma^{i+}_{j^{\prime}}[t]$, ~~ $\tilde{\rho}^i_{j^{\prime}j}[t] \gets \tilde{\sigma}^{i+}_{j^{\prime}}[t]$\;}
{ $\rho^i_{j^{\prime}j}[t] \gets \rho^i_{j^{\prime}j}[t-1]$, ~~$\tilde{\rho}^i_{j^{\prime}j}[t] \gets \tilde{\rho}^i_{j^{\prime}j}[t-1]$\;}
}
$ z_j^{i+}[t] \gets \frac{z_j^{i}[t-1]}{d_j^i[t]+1} +  \sum_{j^{\prime}\in \calI_j^i[t]} \pth{\rho^i_{j^{\prime}j}[t] - \rho^i_{j^{\prime}j}[t-1]}$\; 

$m_j^{i+}[t]  \gets \frac{m_j^i[t-1]}{d_j^i[t]+1} + \sum_{j^{\prime}\in \calI_j^i[t]}(\tilde{\rho}^i_{j^{\prime}j}[t] -\tilde{\rho}^i_{j^{\prime}j}[t-1])$.

$\sigma^i_j[t]  \gets  \sigma^{i+}_j[t] + \frac{z_j^{i+}[t]}{d_j^i[t]+1}$,
$\tilde{\sigma}^i_j[t]  \gets  \tilde{\sigma}^{i+}_j[t] + \frac{m_j^{i+}[t]}{d_j^i[t]+1}$,
$z_j^i[t]  \gets \frac{z_j^{i+}[t]}{d_j^i[t]+1}$,
$m_j^i[t] \gets \frac{m_j^{i+}[t]}{d_j^i[t]+1}$\; 
}

\If{$j$ is a designated agent of network $S_i$}
{
\If{$t\mod \Gamma =0$}
{
Send $\frac{1}{2}z_j^i[t]$ and $\frac{1}{2}m_j^i[t]$ to the PS\; 

Upon receiving messages from the PS {\bf do} \\
update  
$z_j^{i}[t]\gets \frac{1}{2}z_j^{i}[t] + \frac{1}{2M}\sum_{i=1}^M z_{i_0}^i[t]$\; 
$m_j^{i}[t]\gets \frac{1}{2}m_j^{i}[t] + \frac{1}{2M}\sum_{i=1}^M m_{i_0}^i[t]$\;
}
} 

\If{$t\mod \Gamma =0$}
{
The PS does the following: 

Wait to receive $z_{i_0}^i[t]$ and $m_{i_0}^i[t]$ from each designated agent of the $M$ networks\; 

Compute and send $\frac{1}{M}\sum_{i=1}^M \frac{1}{2}z_{i_0}^i[t]$ and $\frac{1}{M}\sum_{i=1}^M \frac{1}{2}m_{i_0}^i[t]$ to all designated agents $i_0$ for $i=1, \cdots, M$. 
} 

\end{algorithm}
Similar to the standard push-sum \cite{kempe2003gossip}, in addition to the primary variable  $z_j^i$, each agent $j$ keeps a mass variable $m_j^i$ to correct the possible bias caused by the graph structure and uses the ratio $z_j^i/m_j^i$ to estimate the average consensus. The correctness of push-sum relies crucially on mass preservation (i.e., $\sum_{i=1}^M\sum_{j=1}^{n_i} m_j^i[t] = N$) holds for all $t$. The variables $\sigma$, $\tilde{\sigma}$, $\rho$, and $\tilde{\rho}$ are introduced to recover the dropped messages. 
Specifically, $\sigma_j^i[t]$ and $\tilde{\sigma}_j^i[t]$ are used to record how much value and mass that agent $j$ (in subnetwork $i$) have been sent to each of the outgoing neighbors of agent $j$ up to time $t$. Corresponding, $\rho^i_{j^{\prime}j}[t]$ and $\tilde{\rho}^i_{j^{\prime}j}[t]$ are used to record how much value and mass have been received by agent $j$ through the link $(j^{\prime}j)\in \calE_i$. 
On the technical side, we use augmented graphs (detailed in Definition \ref{def: augmented graph} of Appendix \ref{app: consensus}) to show convergence. To control the trajectory smoothness of the $z_j^i/m_j^i$ (at both normal agents and virtual agents), in each iteration, both $z$ and $m$ are updated twice in lines 11 to 13.  

For each network, we choose an arbitrary agent as the network representative, and only this designated agent will exchange messages with the PS. Let $i_0$ denote the designated agent of network $i$. 
For every other $\Gamma$ iteration, each designated agent pushes 1/2 of its local value and mass to the PS. The PS computes the received average value and mass and sends the averages back to each designated agent. Each designated agent then updates its local value and mass as once pushed back from the PS.  

\begin{assumption}
\label{ass: connectivity}
Each sub-network $(\calV_i, \calE_i)$ is strongly connected for $i=1, \cdots, M$. 
\end{assumption}

\vspace{0.2em}
Assumption \ref{ass: connectivity} is quite natural and easy to satisfy. In particular, we can run algorithms such as variants of DFS to find the strongly connected components of the original network; see \cite[Chapter 22.5]{cormen2022introduction} for detailed procedures.

\vspace{0.5em}
Denote the diameter of $G(\calV_i, \calE_i)$ as $D_i$. Let $D^* := \max_{i\in [M]}D_i$.  
Let $\beta_i= \frac{1}{\max_{j\in \calV_i} (d_j^{i}+1)^2}$. 
\begin{theorem}
\label{rps convergence rate}
Choose $\Gamma = BD^*$.  
Suppose that Assumptions \ref{ass: link reliability} and \ref{ass: connectivity} hold, and that $t\ge 2\Gamma$.
Then 
\begin{align*}
\norm{\frac{z_j^i[t]}{m_j^i[t]} -\frac{1}{N}\sum_{i=1}^{M}\sum_{j=1}^{n_i} w_j^i} 
&\le \frac{4M^2\sum_{i=1}^{M}\sum_{j=1}^{n_i}\norm{w_{j}^i} \gamma^{\lfloor \frac{t}{2\Gamma}\rfloor - 1}}{\pth{\min_{i\in [M]}\beta_i}^{2D^*B}N}, 
\end{align*} 
where $\gamma = 1-\frac{1}{4M^2} \pth{\min_{i\in [M]}\beta_i}^{2D^*B}$. 
\end{theorem}
Henceforth, for ease of exposition, we adopt the simplification that $\lfloor t/2\Gamma \rfloor - \lceil r/2\Gamma \rceil =  (t-r)/2\Gamma.$ 
Such simplification does not affect the order of convergence rate. 
The exact expression can be recovered while straightforward bookkeeping of the floor and ceiling in the calculation. 

Theorem \ref{rps convergence rate} says that, despite packet-dropping link failures and sparse communication between the networks and the PS, the consensus error $\norm{\frac{z_j^i[t]}{m_j^i[t]} -\frac{1}{N}\sum_{i=1}^{M}\sum_{j=1}^{n_i} w_j^i}$ decays to 0 exponentially fast.  
The more reliable the network (i.e. smaller $B$) and the more frequent across networks information fusion (i.e. smaller $\Gamma$), the faster the convergence rate. 
\begin{remark}
Partitioning the agents into $M$ subnetworks immediately leads to smaller network diameters $D^*$. 
Hence, compared with a gigantic single network, the term $\pth{\min_{i\in [M]}\beta_i}^{2D^*B}$ for the $M$ sub-networks is significantly larger, i.e., faster convergence.  
\end{remark}

\begin{remark}
It turns out that our bound in Theorem \ref{rps convergence rate} is loose in quantifying the total number of global communications. Specifically, for any given $\epsilon>0$, to reduce the error to $O(\epsilon)$, based on the bound in Theorem \ref{rps convergence rate}, it takes 
$t\ge \Omega\pth{\Gamma \log \epsilon/\log \gamma}$, i.e., larger $\Gamma$ leads to slower convergence to $O(\epsilon)$.  
However, our preliminary simulation results (presented in Fig.\,\ref{fig:small_convergence} and Fig.\,\ref{fig:large_convergence}) indicate that if the cost of global communication is significant enough (i.e., takes longer than agent-agent communication within a subnetwork), then a larger $\Gamma$ may sometimes lead to faster convergence in terms of total wall clock time delay.
\end{remark}

\section{State Estimation}
\label{sec: state estimation}
In this section, we study the special case of Eq.\eqref{eq: truth dynamics} when $A=I$, i.e., the state estimation problem. 
We use a ``consensus'' + ``innovation'' approach with Algorithm \ref{alg:push-sum hierarchical FL} as the consensus component and
use dual averaging as the innovation component. 
We add the following lines of pseudo-code right after line 12 inside the outer {\bf for}-loop in Algorithm 1 – at the end of the for-loop: 
\vskip 0.6\baselineskip
\fbox{\begin{minipage}{22em}
{\em 
Obtain observation $y_j^i[t]$; \\
Compute a stochastic gradient $g_j^i[t]$ of $f_{j}^i$; \\
$z_j^i[t]\gets z_j^i[t-1] + g_j^i[t]$; \\
$w_j^i[t] \gets\prod_{w\in \calW}^{\varphi}(\frac{z^i_j[t]}{m^i_j[t]}, \eta[t-1])$. 
} 
\end{minipage}}
\vskip 0.6\baselineskip
Next, we explain the added pseudo-code.
\begin{enumerate}[(I)]
    \item Each agent $j$ first obtains a new observation $y_i[t]$ according to Eq.\,\eqref{eqn: local observation}.  

    \item For ease of exposition, the local function $f_j^i$ at agent $j$ of network $S_i$ is defined as:
 \begin{equation} \label{eqn: local asymptotic function}
    f^i_j(w):= \frac{1}{2} \mathbb{E}[\| H^i_jw - y^i_j\|_2^2], 
 \end{equation} 
 where $w\in \calW$. 
 Let $F$ denote the global objective, i.e.,  
 \begin{align}
 \label{eq: global objective: state}
 f :=\sum_{i=1}^M \sum_{j=1}^{n_i} f^i_j. 
 \end{align}
Despite the fact that $f_j^i$ is well-defined, it is unknown to agent $j$. This is because the observation noise $\xi_j^i$ distribution is unknown and agent $j$ cannot evaluate the expectation in \eqref{eqn: local asymptotic function}. Hence, we cannot perform the standard distributed dual averaging. 
Fortunately, as $H_j^i$ is known, agent $j$ can access the natural stochastic gradient of \eqref{eqn: local asymptotic function}  
 \begin{equation}
\label{eq: local client stochastic gradient}
g^i_j[t] = H_j^{\top}\pth{H_j w[t-1]-y^i_j[t-1]}.  
 \end{equation}
We calculate $g^i_j[t]$ with $y^i_j[t-1]$ instead of $y^i_j[t]$ is for ease of exposition in the analysis. We can replace $y^i_j[t-1]$ with $y^i_j[t]$, the analysis remains the same except for time index changes.

\item The variable $z_j^i[t]$ is used to cumulative all the locally computed stochastic gradients up to time $t$.  

\item The update of the local estimate $w_j^i$ uses the function 
\begin{equation}
    \prod_{w\in \calW}^{\varphi} \pth{z, \eta} := \arg\min_{w\in \calW}\sth{\iprod{z}{w} + \frac{1}{\eta}\varphi(w)}
\end{equation}
where $\eta>0$ is the stepsize, and $\varphi: \reals^d \to \reals$ is a non-negative and 
$1$-strongly convex function with respect to $\ell_2$ norm, i.e., 
$\varphi (w^{\prime}) \ge \varphi (w) + \iprod{\nabla \varphi (w)}{w^{\prime} - w} + \frac{1}{2} \norm{w^{\prime} - w}$.  
One example of such $1$-strongly convex function is the $\ell_2$ norm, i.e., $\varphi(w) = \frac{1}{2}\norm{w}^2$. 
We will choose a sequence of delaying stepsizes $\{\eta[t]\}_{t=0}^{\infty}$, which will be specified in Theorem \ref{thm: convergence of hierarchical FL}. 
\end{enumerate}

\begin{assumption}
\label{ass: compact}
The constraint set $\calW$ is compact. 
\end{assumption}

In many real-world applications such as power grids and unmanned aerial vehicles, the constraint set is bounded and closed. Dealing with constraint learning is often deemed as more challenging yet practical than unconstrained ones \cite{nesterov2018lectures}. 
\begin{assumption}
\label{ass: global observability}
Let $K\triangleq \sum_{i=1}^M\sum_{j=1}^{n_i}\pth{(H_j^i)^{\top}H_j^i}$. 
The matrix $K$ is positive definite.  
\end{assumption}

It is easy to see that Assumption \ref{ass: global observability} is necessary even for the single network (i.e., $M=1$), failure-free (i.e., $B=1$), and noiseless (i.e., $\xi_j^i[t]=0$) settings.

Denote the diameter of set $\calW$ as 
 \begin{equation}
\label{eq: diameter}    
R_0:=\max_{w, w^{\prime}\in \calW} \norm{w - w^{\prime}} 
 \end{equation}  
\begin{proposition}
\label{prop: lipschitz}
Suppose that Assumption \ref{ass: compact} holds. Then $f$ is $L$-Lipschitz continuous with $L:=R_0\norm{K}$, i.e., $\norm{f(w) - f(w^{\prime})} \le  L\norm{w-w^{\prime}}$ for all $w, w^{\prime}\in \calW.$ 
Moreover, $f_j^i$ is also $L$-Lipschitz continuous $\forall i\in [M], j\in \calV_i$. 
\end{proposition}

\begin{assumption}
\label{ass: bounded observation noise}
The observation noise $\xi_j^i[t]$ is independent across time $t$ and across agents $j\in \calV_i$ for $i=1, \cdots, M$. Moreover, $\norm{H_{ij}^{\top}\xi_j^i}\le B_0$ for all agents.  
\end{assumption}

Similar assumptions are adopted in \cite{kar2008distributed,kar2012distributed,su2019finite}.
Under Assumptions \ref{ass: compact} and \ref{ass: bounded observation noise}, the following proposition holds immediately. 
\begin{proposition}
 \label{prop: bounded stochastic gradient}
 Suppose that Assumptions \ref{ass: compact} and \ref{ass: bounded observation noise} hold. 
Let $L_0:= 2R_0\norm{K} + B_0$. It is true that $\norm{g_{j}^i(w)}\le L_0$.  
\end{proposition}
\vskip 0.2\baselineskip
For each agent, we define $\hat{w}_j^{i}[t] := \frac{1}{t}\sum_{r=1}^t w_j^{i}[r]$ to be the running average $w_j^{i}[t]$. 
\begin{theorem}
\label{thm: convergence of hierarchical FL}
Choose $\Gamma = BD^*$. 
Suppose that Assumptions \ref{ass: link reliability}-\ref{ass: global observability} hold, $t\ge 2\Gamma$, and that $\varphi(w^*)\le R^2$.
Let $\sth{\eta[t]}_{t=0}^{\infty}$ be a sequence of non-increasing step sizes. 
For any $\delta\in (0, 1)$,   
the following holds with probability at least $(1-\delta)$:  

\begin{align*}
&\norm{\hat{w}_j^{i}[t]-w^{*}}^2 \le \frac{1}{\lambda_{\min}} \left(\frac{NL_0^2}{2t}\sum_{r=1}^t \eta[r-1]  + \frac{NR^2}{t\eta[t]} \right.\\
& + \left.\frac{ 4M^2L_0^2 \gamma^{\frac{1}{2\Gamma}}}{\pth{1-\gamma^{\frac{1}{2\Gamma}}}\pth{\min_{i\in [M]}\beta_i}^{2D^*B}} \frac{1}{t}\sum_{r=1}^t \eta[r-1] + 4NLR\sqrt{\frac{\log \frac{1}{\delta}}{t}}.\right)      
\end{align*}
where $\lambda_{\min}$ is the smallest eigenvalue of $K$. 
\end{theorem}

\begin{remark}
Choosing $\eta[t] = \frac{1}{\sqrt{t}}$ 
for $t\ge 1$ with $\eta[0]=1$
, it becomes 
\begin{equation*}
    \norm{\hat{w}_j^{i}[t]-w^{*}}^2 \le \frac{1}{\lambda_{\min}} \left(\frac{NL_0^2}{2\sqrt{t}}  + \frac{N}{ \sqrt{t}}R^2 
+ \frac{ 4M^2 L_0^2 \gamma^{\frac{1}{2\Gamma}}}{\pth{1-\gamma^{\frac{1}{2\Gamma}}}\pth{\min_{i\in [M]}\beta_i}^{2D^*B}} \frac{1}{\sqrt{t}} + 4NLR\sqrt{\frac{\log \frac{1}{\delta}}{t}}\right)    
\end{equation*}
\end{remark}

\section{State Tracking}
\label{sec: tracking}
In the state tracking problem, the agents try to collaboratively track $w^*[t]$. We present the full description of our algorithm in Algorithm \ref{alg: tracking single}. Our algorithm uses projected gradient descent as the local innovation component.

In each round, in line 4, each agent gets a new observation $y_j^i[t]$. However, the local stochastic gradient is computed on the measurement obtained in the previous round $t-1$. As can be seen from our analysis, we use such one step setback to align the impacts of the global dynamics $A$ with the relevant parameters evolution. 
In addition,  in line 18, we apply $A$ to the local $z$ sequence update.
Recall that if a link does not function properly, the sent value and mass are stored in virtual nodes (the nodes that correspond to the edges). Hence, we apply $A$ to the auxiliary variables $\sigma$ and $\rho$ as well. Specifically, we update $\rho_{j^{\prime}j}^i$ and $\tilde{\rho}_{j^{\prime}j}^i$ twice -- the first time in lines 9-12, and the second time in lines 16 and 17. 
In line 15, we apply $A$ to the original update of $\sigma$. Notably, we apply $A$ to $z$ and auxiliary variables that are relevant to $z$ only; we do not apply $A$ to the mass update. 
In line 19, $\prod_{\calW}[\cdot]$ is an operator that projects any given $w\in \reals^d$ onto $\calW$. This projection is used to ensure the boundedness of stochastic gradients.

\begin{algorithm}
\caption{Collaborative Tracking Algorithm}
\label{alg: tracking single}
 {\em Initialization}: 
For each sub-network $i=1, \cdots, M$: $z_j^i[0]=w_j^i\in \reals^d$, $m^i_j[0]=1\in \reals,$ 
$\sigma_j^i[0]={\bf 0}\in \reals^d$, $\tilde{\sigma}_j^i[0]=0\in \reals$, and $\rho^i_{j^{\prime}j}[0]={\bf 0}\in \reals^d$, $\tilde{\rho}^i_{j^{\prime}j}[0]=0\in \reals$ for each incoming link, i.e., $j^{\prime} \in \calI_j^i$.

\vskip 0.2\baselineskip 
In parallel, each agent in parallel does:\\
\For{$t\ge 1$}
{ 
Obtain measurement $y^i_j[t]$\;   
Compute a local stochastic gradient $g^i_j[t]$ that corresponds to the previous measurement $y^i_j[t-1]$ evaluated at local estimated $w^i_j[t-1]$\;  

$\sigma_j^{i+}[t]  \gets  \sigma_j^i[t-1] + \frac{z_j^i[t-1]}{d_j^i[t]+1}$,
$\tilde{\sigma}_j^{i+}[t] \gets \tilde{\sigma}_j^i[t-1] + \frac{m_j^i[t-1]}{d_j^i[t]+1}$\;

Broadcast $\pth{\sigma^{i+}_j[t], \tilde{\sigma}^{i+}_j[t]}$ to outgoing neighbors\;

\For {each incoming link $(j^{\prime},j)\in \calE_i[t]$}
{\eIf{message $\pth{\sigma^{i+}_{j^{\prime}}[t], \tilde{\sigma}^{i+}_{j^{\prime}}[t]}$ is received}
{$\rho^{i+}_{j^{\prime}j}[t] \gets \sigma^{i+}_{j^{\prime}}[t]$, ~~ $\tilde{\rho}^{i+}_{j^{\prime}j}[t] \gets \tilde{\sigma}^{i+}_{j^{\prime}}[t]$\;}
{ $\rho^{i+}_{j^{\prime}j}[t] \gets \rho^i_{j^{\prime}j}[t-1]$, ~~$\tilde{\rho}^{i+}_{j^{\prime}j}[t] \gets \tilde{\rho}^i_{j^{\prime}j}[t-1]$\;}
}
$ z_j^{i+}[t] \gets \frac{z_j^{i}[t-1]}{d_j^i[t]+1} +  \sum_{j^{\prime}\in \calI_j^i[t]} \pth{\rho^i_{j^{\prime}j}[t] - \rho^i_{j^{\prime}j}[t-1]}$\;  

$m_j^{i+}[t]  \gets \frac{m_j^i[t-1]}{d_j^i[t]+1} + \sum_{j^{\prime}\in \calI_j^i[t]}(\tilde{\rho}^i_{j^{\prime}j}[t] -\tilde{\rho}^i_{j^{\prime}j}[t-1])$\; 

$\sigma^i_j[t]  \gets  A(\sigma^{i+}_j[t] + \frac{z_j^{i+}[t]}{d_j^i[t]+1})$, $\tilde{\sigma}^i_j[t]  \gets  \tilde{\sigma}^{i+}_j[t] + \frac{m_j^{i+}[t]}{d_j^i[t]+1}$\; 

\For{each incoming link $(j^{\prime},j)\in \calE_i[t]$}
{
$\rho_{j^{\prime}j}^i[t] \gets A \rho_{j^{\prime}j}^{i+}[t]$, \text{and} 
$\tilde{\rho}_{j^{\prime}j}^i[t] \gets  \tilde{\rho}_{j^{\prime}j}^{i+}[t]$. 

}

$z_j[t] \gets A\frac{z_j^+[t]}{d_j[t]+1} - \eta[t]g_j[t]$, and $m_j[t] \gets \frac{m_j^+[t]}{d_j[t]+1}$\; 
$w_j[t] = \prod_{\calW}\qth{\frac{z_j[t]}{m_j[t]}}$.  
}

\If{$j$ is a designated agent of network $S_i$}
{
\If{$t\mod \Gamma =0$}
{
Send $\frac{1}{2}z_j^i[t]$ and $\frac{1}{2}m_j^i[t]$ to the PS\; 

Upon receiving messages from the PS {\bf do} \\
update  
$z_j^{i}[t]\gets \frac{1}{2}z_j^{i}[t] + \frac{1}{2M}\sum_{i=1}^M z_{i_0}^i[t]$\; 
$m_j^{i}[t]\gets \frac{1}{2}m_j^{i}[t] + \frac{1}{2M}\sum_{i=1}^M m_{i_0}^i[t]$\;
}
} 

\If{$t\mod \Gamma =0$}
{
The PS does the following: 

Wait to receive $z_{i_0}^i[t]$ and $m_{i_0}^i[t]$ from each designated agent of the $M$ networks\; 

Compute and send $\frac{1}{M}\sum_{i=1}^M \frac{1}{2}z_{i_0}^i[t]$ and $\frac{1}{M}\sum_{i=1}^M \frac{1}{2}m_{i_0}^i[t]$ to all designated agents $i_0$ for $i=1, \cdots, M$. 
}
\end{algorithm}

\vskip \baselineskip
In addition to the aforementioned assumptions, the following assumptions will be used in our analysis. 
\begin{assumption}
\label{ass: tracking: consensus}
The global linear dynamic matrix $A$ is positive semi-definite with $\norm{A}\le 1$.  
\end{assumption}
Intuitively, Assumption \ref{ass: tracking: consensus} ensures that the underlying truth $w^*[t]$ is within a bounded region.  It's worth noting that $\norm{\bm{I}}=1$, thereby fulfilling Assumption \ref{ass: tracking: consensus}. 
 
Let $\bar{z}[t] = -\frac{1}{N}\sum_{i=1}^M \sum_{j=1}^{n_i} z_j^{i}[t]$, which differs from standard aggregation by a ``-'' sign. 
Recall that $\gamma = 1-\frac{1}{4M^2} \pth{\min_{i\in [M]}\beta_i}^{2D^*B}$ as per Theorem \ref{rps convergence rate}. The following holds.  
\begin{theorem}
\label{thm: convergence of hierarchical FL: tracking}
Suppose that Assumptions \ref{ass: link reliability}-\ref{ass: tracking: consensus} hold. 
Choose $\Gamma = BD^*$, $\eta[t] = \frac{1}{\lambda_1 t}$ for $t\ge 1$ with $\eta[0]=\frac{1}{\lambda_1}$.  
Let $b=\norm{A} \gamma^{\frac{1}{\Gamma}}$, $t_0 = \frac{2}{\log 1/b} \log \pth{\frac{2}{\log 1/b}}$, and $\bar{t}_0 = \max\{t_0, 2\Gamma\}$.  
Then 
\begin{equation} 
\label{eq: bound of term b}
\norm{w_j^i[t] - \bar{z}[t]} \le 
\begin{cases}
\frac{16M^2L_0}{\pth{\min_{i\in [M]}\beta_i}^{2D^*B}\lambda_1 (1-b)t}, &\text{if } ~ t \ge \bar{t}_0\\ 
\frac{4M^2L_0 t_0}{\pth{\min_{i\in [M]}\beta_i}^{2D^*B}\lambda_1},  &\text{if } ~ t < \bar{t}_0
\end{cases}
\end{equation}
Moreover, when $t\ge \bar{t}_0$, for any given $\delta\in (0,1)$, the following holds with probability at least $(1-\delta)$: 
\begin{align}
\label{eq: convergence tracking}
&  \norm{w^i_j[t] - w^*[t]}  \le \norm{w_{0}^*} \exp\pth{\frac{\lambda_1}{\lambda_d}}\frac{1}{t} + \frac{\exp\pth{\lambda_1/\lambda_d}4M^2L_0 t^2_0}{\pth{\min_{i\in [M]}\beta_i}^{2D^*B}\lambda_1}\frac{1}{t} \nonumber\\
&  \qquad + \frac{32M^2L_0\exp\pth{\lambda_1/\lambda_d}}{\pth{\min_{i\in [M]}\beta_i}^{2D^*B}\lambda_1(1-b)}\frac{\log(t+1)}{t} + \frac{2B_0\exp\pth{\lambda_1/\lambda_d}}{\lambda_1}\sqrt{\frac{d}{2t} \log (d/\delta)},      
\end{align}
where $\lambda_1 \ge \cdots \ge \lambda_d = \lambda_{\min}>0$ are the eigenvalues of $K$. 
\end{theorem}
\begin{remark}
For sufficiently large $t$, the dominance term in the upper bound of Eq.\eqref{eq: convergence tracking} is \\
$\sqrt{\frac{d}{2t}\log (d/\delta)}B_0\log t$, which arises from the observation noise.          
\end{remark}

\input{experiments_v2.tex}
\input{conclusion.tex}

\section*{Acknowledgements}
This research was supported by ONR award N00014-18-9-0001.

\newpage

\bibliographystyle{acm} 
\bibliography{citation} 
\newpage

\appendix
\input{appendix.tex}

\end{document}

%% file: experiments_v2.tex
\section{Numerical Results}
\label{sec: numericals}

\subsection{System and Threat Models}
We have conducted simulations in two different environments: a small-scale network inspired by underwater acoustic (UWA) networks, and a large-scale synthetic network. 

\begin{figure}[t!]
    \centering
    \begin{subfigure}[t]{0.45\textwidth}
        \centering
       \includegraphics[height=5cm]{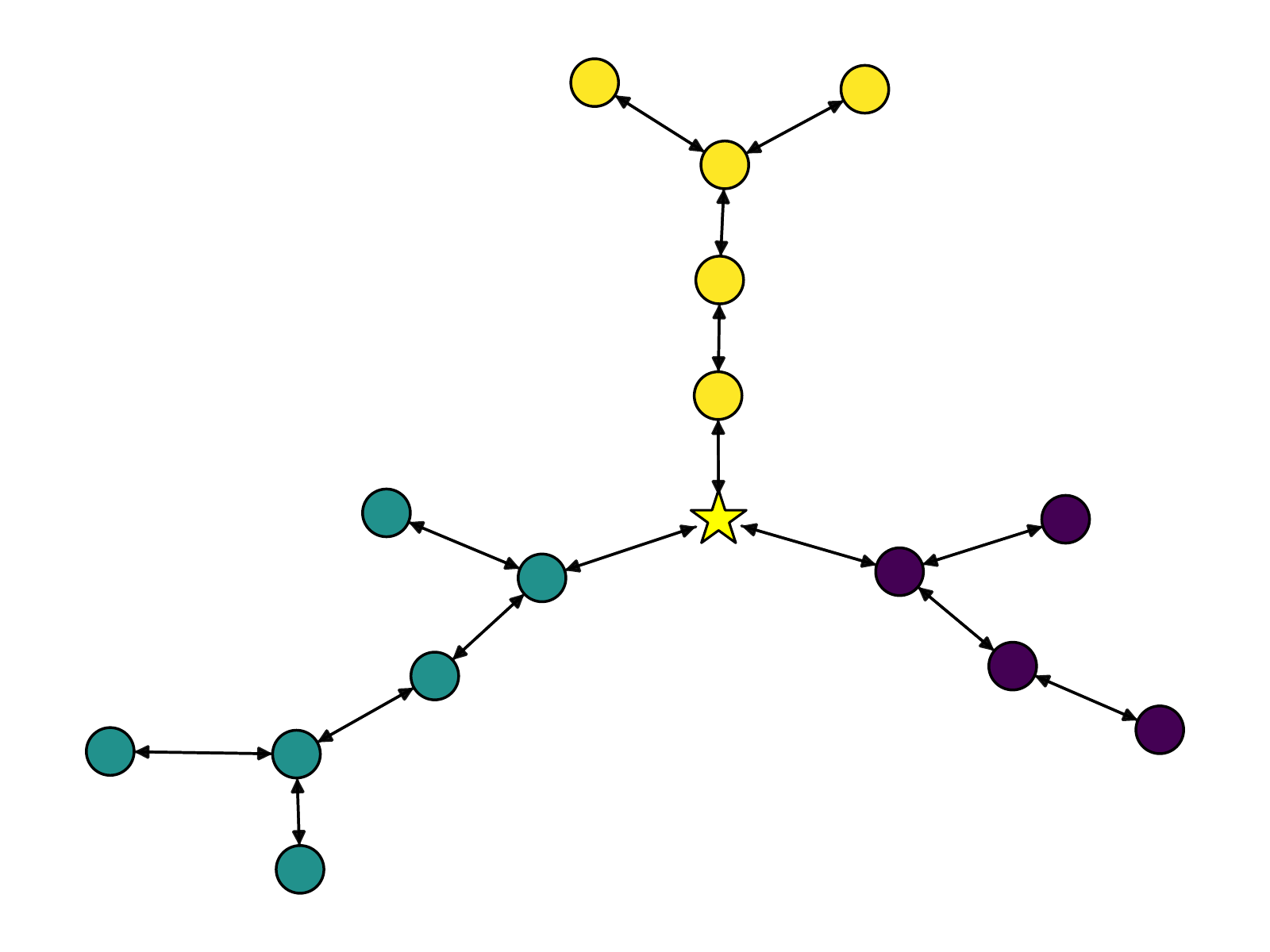}
       \caption{Small Network Structure}
       \label{fig:small_network} 
    \end{subfigure}%
    ~
    \begin{subfigure}[t]{0.45\textwidth}
        \centering
       \includegraphics[height=5cm]{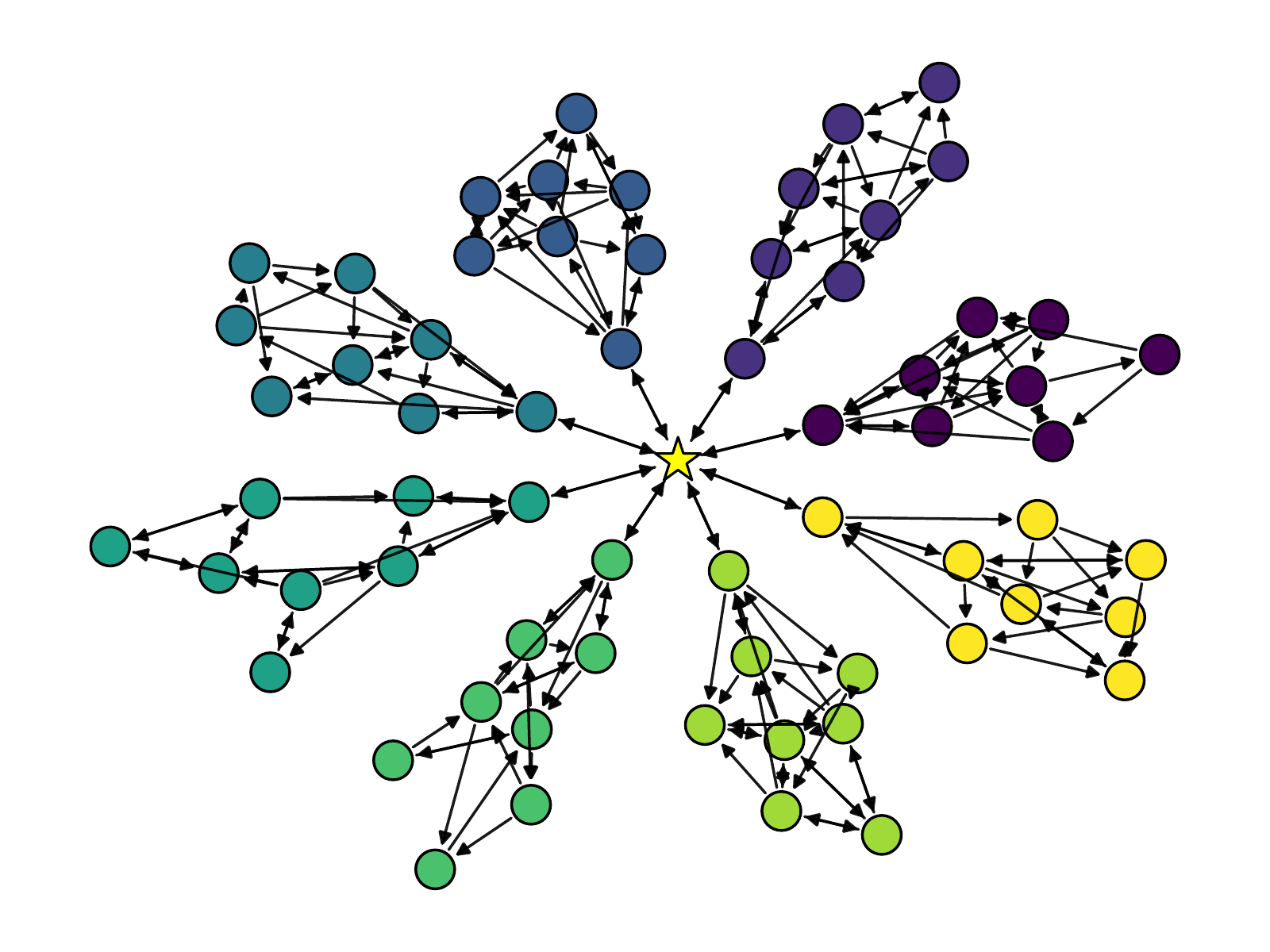}
       \caption{Large Network Structure}
       \label{fig:large_network}
    \end{subfigure}
    \caption{Network Configurations}
\end{figure}

\paragraph{Underwater Acoustic Networks}
For the case of underwater acoustic networks, we consider a 16-node network with three clusters and one parameter server node, as shown in Figure \ref{fig:small_network}. We consider each set of nodes branching off of the parameter server, denoted by a star, to be a cluster, similar to \cite{underwater_acoustic}. We use the AUVNetsim library \cite{auvnetsim} to simulate the network infrastructure. Notably, each node generates packets according to a Poisson distribution where the average rate $\lambda$ doubles when communicating with the parameter server. In this setting, packet-dropping links are simulated by signal collisions encountered in acoustic networks. 

\paragraph{A Larger-Scale Synthetic Network}
We generated a synthetic 64-node graph with 8 clusters according to the stochastic block model, as shown in Figure \ref{fig:large_network}. Specifically, we regenerated each cluster with an inter-node edge probability of 0.3 until it was strongly connected. We considered all edges to have an equal communication cost $\lambda$. Communication with the parameter server incurred an additional cost of $2\lambda$, as in our 16-node network. We introduce packet-dropping links by setting the delay between rounds of edge availability as the minimum of variable $B$ and a geometric distribution with parameter $p=\frac{1}{1.5B}$.

\subsection{State and Observation Models}
We consider two separate observation models for state estimation and tracking, respectively. 

For state estimation, our ground truth state is a drawn from a $d$-dimensional zero-mean Gaussian with identity covariance, where $d=9$ for the small network and $d=25$ for the large network.  
We randomly sample observation matrices for each client such that no single cluster can observe the entire state, but the state is fully observable across all clients.

For state tracking, we use a 2-dimensional Gaussian vector representing X and Y coordinates of an object moving according to global dynamics $A = aI$, with $a=0.99$. Each agent has an identity observation matrix and i.i.d. Gaussian noise with variance drawn from $U(0, 2\bar{\sigma})$ added to each dimension of the observation. We use $\bar{\sigma}=0.2$ in our experiments unless specified otherwise.

\subsection{Method Evaluation}
We compare our hierarchical decomposition to a single-network approach on an augmented graph generated as follows: rather than inter-network communication going through the parameter server, we add a bidirectional edge between random members of each pair of clusters to facilitate inter-group communication. Communication using these added edges incurs a delay of $2\lambda$, similar to using the parameter server. We run all methods with an initial step size of 0.01 for state estimation and 0.1 for tracking. We monitor the average L2 state estimation error across agents as a function of the cumulative system delay (in terms of $\lambda$).

\begin{figure}[h]
    \centering
    \includegraphics[width=0.48\textwidth]{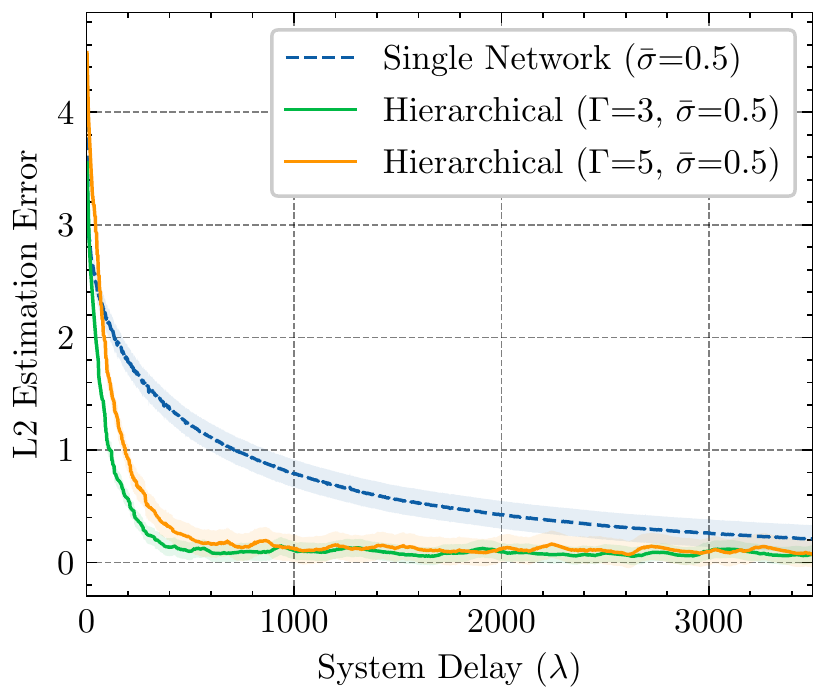}
    \caption{State Estimation on Small Network Configuration.}
    \label{fig:small_convergence}
\end{figure}

We first consider the benefit of our hierarchical approach in the smaller UWA network environment. In Figure \ref{fig:small_convergence}, we plot the average L2 state estimation error against the cumulative system delay for various system configurations. It is clear that the hierarchical methods reach a low state estimation error much faster than the single-network approach, requiring less than a third of the total time $\lambda$ to obtain near zero error. In this small-scale setting, the choice of synchronization frequency $\Gamma$ has minimal effect on the performance of our method.

\begin{figure}[!ht]
    \centering
    \includegraphics[width=0.48\textwidth]{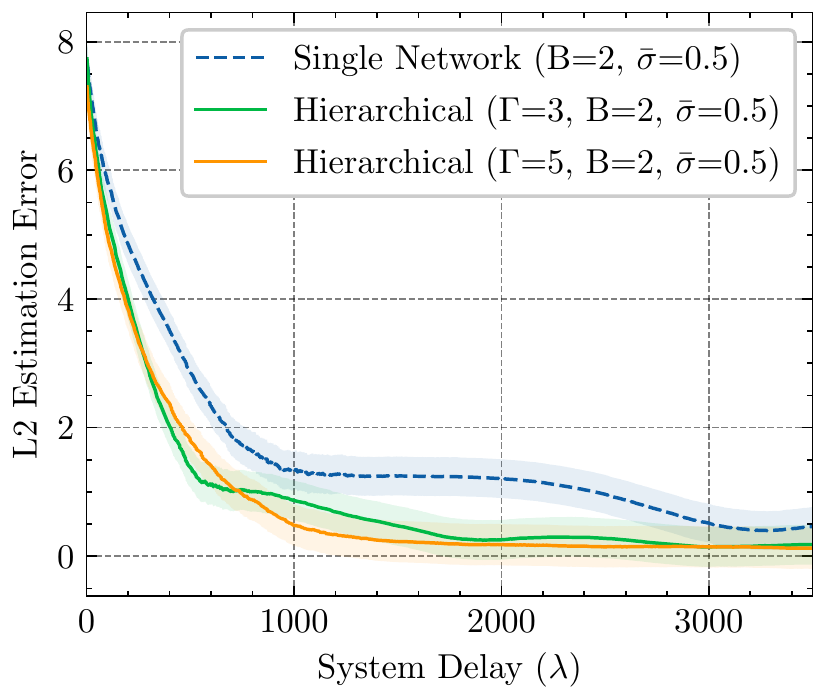}
    \caption{Comparison of Synchronization Frequency on Large Network Configuration.}
    \label{fig:large_convergence}
\end{figure}

\begin{figure}[!ht]
    \centering
    \includegraphics[width=0.48\textwidth]{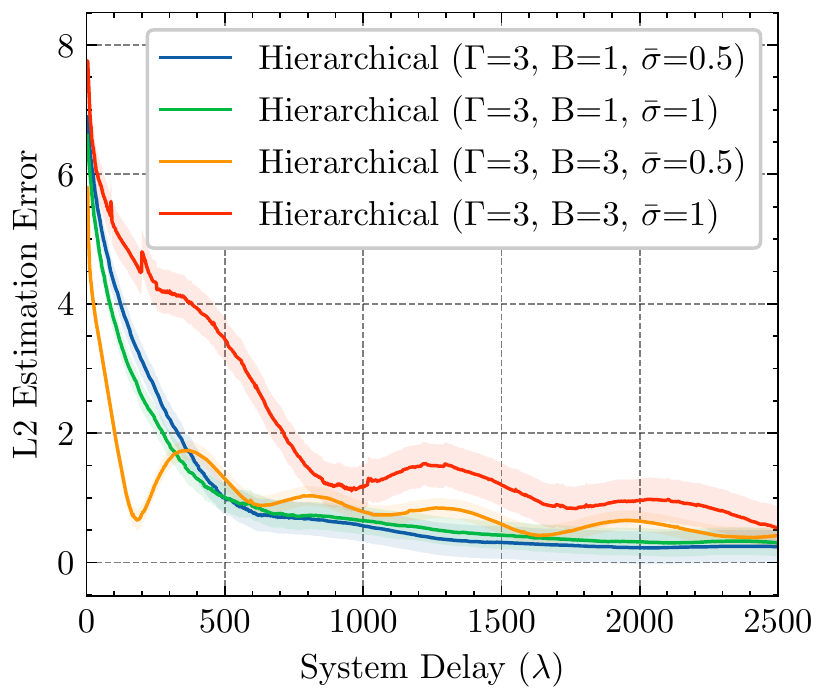}
    \caption{Ablation on System Robustness ($B, \bar{\sigma}$) on Large Network Configuration.}
    \label{fig:robustness}
\end{figure}

We conduct a similar experiment on the large system setting with fixed link availability $B$ and show the results in Figure \ref{fig:large_convergence}. In this setting we see a clear benefit of favoring inexpensive within-network communication, with the higher $\Gamma$ configuration exhibiting the fastest convergence. We additionally conduct an ablation study on the effect of link drops and observation noise in Figure \ref{fig:robustness}.

\begin{figure}[!ht]
    \centering
    \includegraphics[width=0.48\textwidth]{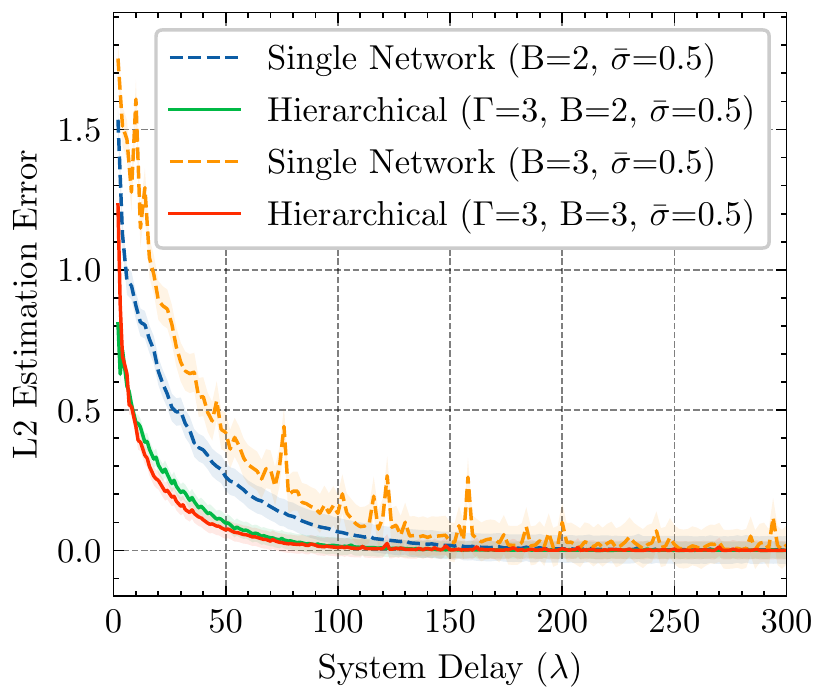}
    \caption{Comparison of Methods for State Tracking.}
    \label{fig:tracking}
\end{figure}

\begin{figure}[!ht]
    \centering
    \includegraphics[width=0.48\textwidth]{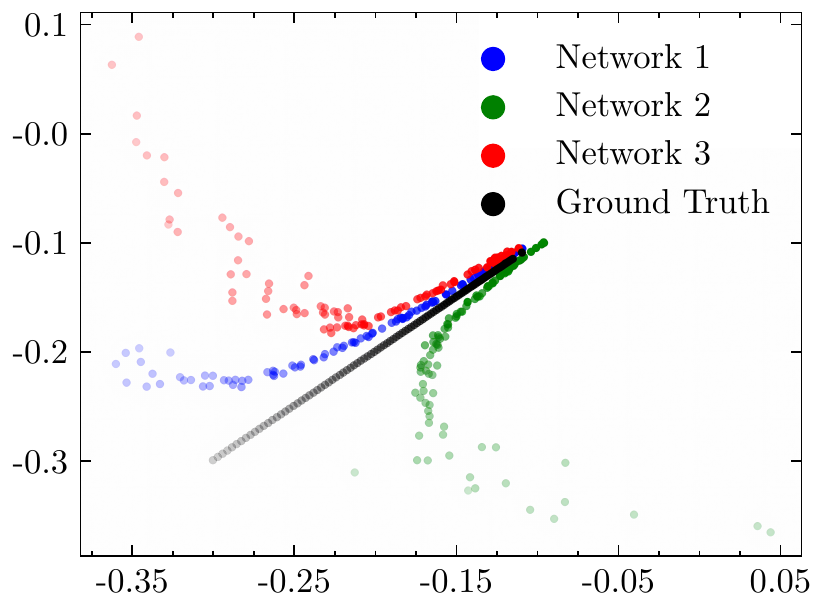}
    \caption{State Tracking Visualization.}
    \label{fig:trajectory}
\end{figure}

For state tracking, we provide another comparison of our hierarchical approach and a single-network approach for two levels of link-drops in Figure \ref{fig:tracking}. We find that our hierarchical method not only obtains a lower estimation error at all points in time, but also is more robust to noise caused by dropping-links. We additionally provide a visualization of the state estimate over time in Figure \ref{fig:trajectory}. Each point represents the average sub-network state estimate, where darker points indicate more recent positions in time. 

%% file: conclusion.tex
\section{Conclusion}
\label{sec: conclusion}
In this article, we have studied the problem of distributed state estimation and tracking under unreliable networks. We designed one algorithm for each setting, each based on a novel hierarchical push-sum routine for improved communication efficiency and network resilience. Compared to existing works, our method makes limited assumptions on the network structure and link failures, and we have provided an analysis of convergence rates to shed light on the balance between network structure, link availability, and our hierarchical synchronization frequency. We demonstrated the practicality of our method through simulation studies considering underwater acoustic networks and large-scale synthetic networks. A key observation is that even in small-scale environments with limited inter-cluster communication delays (i.e., a sensor network spanning a limited geographic region), we still observe improved convergence with our hierarchical decomposition compared to a baseline single-network approach. Future work includes developing techniques for state tracking under nonlinear dynamics and methods for handling adversarial sensor noises. In addition, we would like to learn the optimal clusters under the communication constraints.

%% file: appendix.tex
\section{Hierarchical Consensus}
\label{app: consensus}
\subsection{Matrix Construction}
\label{subec: matrix construction}
\begin{figure}[h]
\centering
\includegraphics[width=0.3\textwidth]{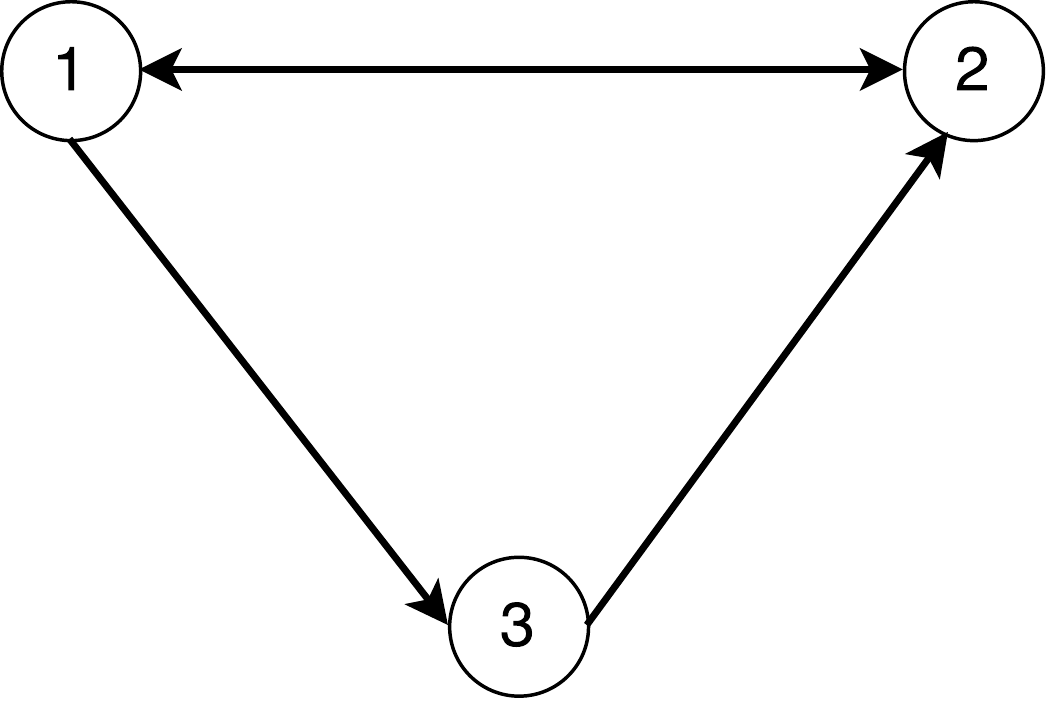}
~~~~~~~~~~~~
\includegraphics[width=0.3\textwidth]{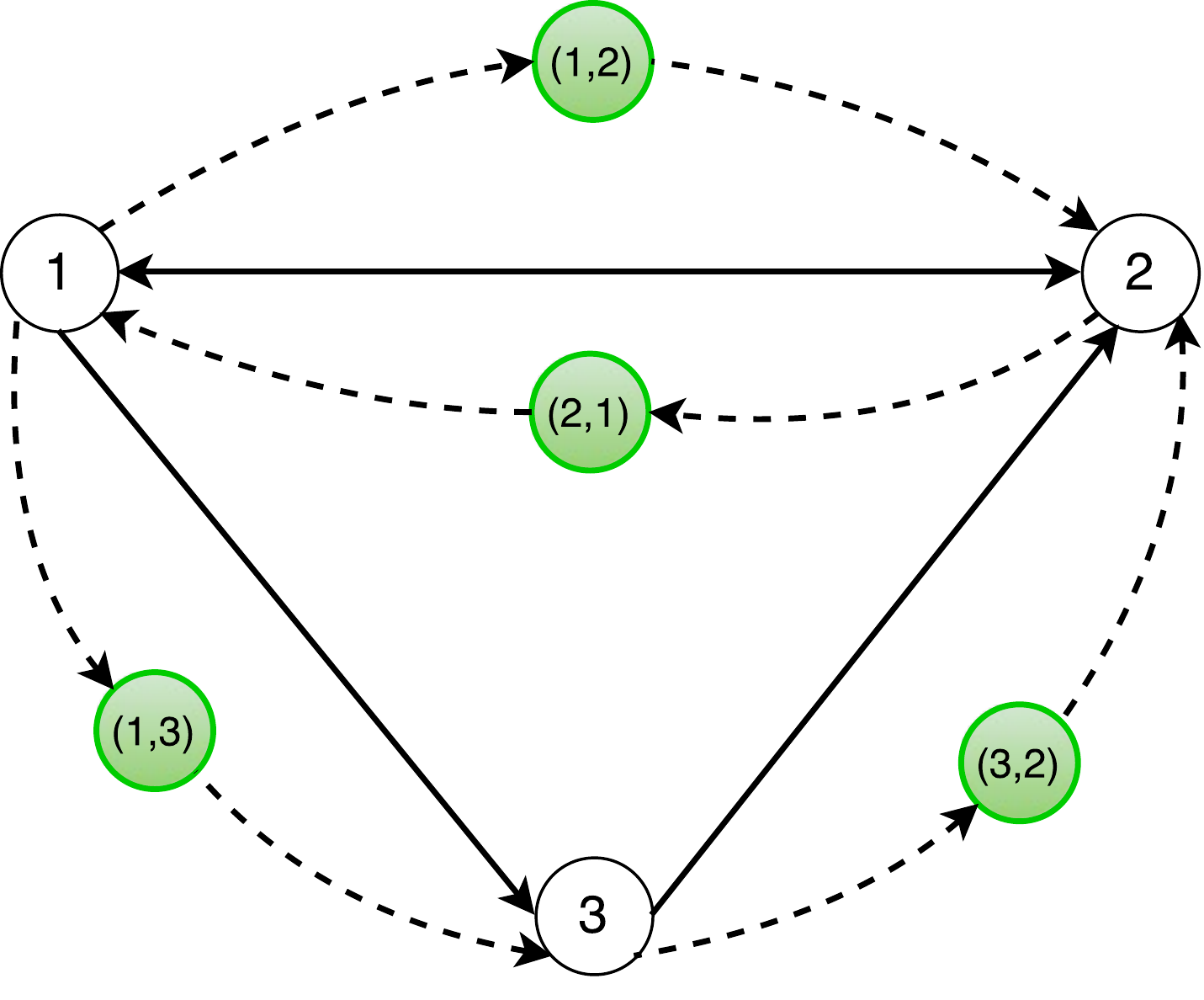}
\caption{Augmented graph example \cite{UnreliableConvergence}}
\label{fig: augmented graph example}
\end{figure}

The analysis of Theorem \ref{rps convergence rate} relies on the notion of augmented graphs and a compact matrix representation of the dynamics of $z$ and $m$ over those augmented graphs.  
\begin{definition}[Augmented Graph] \cite{vaidya2012robust}
\label{def: augmented graph}
Given a graph $G(\calV, \calE)$, the augmented graph $G^a(\calV^a, \calE^a)$ is constructed as: 
\begin{enumerate}
\item $\calV^a=\calV\cup \calE$:  $|\calE|$ virtual agents are introduced, each of which represents a link in $G(\calV, \calE)$. 
Let $n_{j^{\prime}j}$ be the virtual agent corresponding to edge $(j^{\prime},j)$.
\item $\calE^a \triangleq \calE \cup \sth{\pth{j^{\prime}, n_{j^{\prime}j}}, (n_{j^{\prime}j}, j), ~ \forall ~ (j^{\prime}, j)\in \calE}$. 
\end{enumerate}
\end{definition}
An example can be found in Fig.\,\ref{fig: augmented graph example}. In the original graph (left), the node and edge sets are $\calV =\{1, 2, 3\}$ and $\calE=\{(1,2), (2,1), (1,3), (3,2)\}$, respectively. The four green nodes in the corresponding augmented graph (right) are the virtual agents and the dashed arrows indicate the added links.  

We study the information flow in the augmented graphs rather than in the original systems.  
When a message is not successfully delivered over a link, our Algorithm \ref{alg:push-sum hierarchical FL} uses a well-calibrated mechanism to recover such a message and to convert it into a delayed message. Intuitively, we can treat the delayed message as the ones that are first sent to virtual nodes, and then are held for at most $B-1$ iterations, and finally are released to the destination node. 
For each subnetwork $S_i$, let $m_i:= |\calE_i|$ denote the number of edges. Let $\tilde{N}: = \sum_{i=1}^M \pth{n_i + m_i}$. Thus, we construct a matrix $\bm{M}[t] \in \reals^{\tilde{N} \times \tilde{N}}$ as follows. 

\paragraph{Non-global fusion iterations}
Fix a network $S_i$. Fix $t$ be arbitrary iteration such that $t\mod \Gamma \not=0$. The matrix construction is the same as that in \cite{UnreliableConvergence}. For completeness, we present the construction as follows.

For each link $(j,j^{\prime})\in\calE_i[t]$, and $t\ge 1$,
\begin{align}
\sfB^i_{(j,j^{\prime})}[t] \triangleq \left\{
 \begin{tabular}{ll}
 1, ~ if $(j,j^{\prime})\in \calE_i[t]$ and is reliable at time $t$; \\
 0, ~ otherwise.  \label{indicator_var}
 \end{tabular}
 \right.
\end{align}
Recall that $z_j^i$ and $m_j^i$ are the value and mass for $j\in \calV^i$.
For each $(j,j^{\prime})\in \calE_i[t]$,
\begin{align}
z^i_{n_{j^{\prime}j}}[t]\triangleq \sigma^i_{j^{\prime}}[t]-\rho^i_{j^{\prime}j}[t], \quad  
 m^i_{n_{j^{\prime}j}}[t]\triangleq \tilde{\sigma}^i_{j^{\prime}}[t]-\tilde{\rho}^i_{j^{\prime}j}[t], \label{w virtual}
\end{align}
with $z^i_{n_{j^{\prime}j}}[0]={\bf 0}\in \reals^d$ and $m^i_{n_{j^{\prime}j}}[0]=0\in \reals$.
Intuitively, $z^i_{n_{j^{\prime}j}}[t]$ and $m^i_{n_{j^{\prime}j}}[t]$ are the value and weight that agent $j^{\prime}$ tries to send to agent $j$ not not successfully be delivered. 
Let
\begin{align*}
&{\bf M}_{j,j}[t]\triangleq \frac{1}{\pth{d_j^{i}[t]+1}^2}, \\ 
&{\bf M}_{j,j^{\prime}}[t]\triangleq \frac{\sfB^i_{(j^{\prime}, j)}[t]}{\pth{d_j^{i}[t]+1}\pth{d_{j^{\prime}}^i[t]+1}}, ~ \forall ~ j^{\prime}\in \calI_j^i[t],  \\
& {\bf M}_{j, n_{j^{\prime}j}}[t]\triangleq \frac{\sfB^i_{(j^{\prime}, j)}[t]}{d_j^{i}[t]+1}, ~ \forall ~ j^{\prime} \in \calI^i_j[t], \\
&{\bf M}_{n_{j^{\prime}j}, j^{\prime}}[t] \triangleq \frac{1}{\pth{d_{j^{\prime}}^i[t]+1}^2} + \frac{1- \sfB^i_{(j^{\prime}, j)}[t]}{d_{j^{\prime}}^i[t]+1},  \\
& {\bf M}_{n_{j^{\prime}j}, k}[t] = \frac{\sfB^i_{(k,j^{\prime})}[t]}{\pth{d_k^i[t]+1}\pth{d_{j^{\prime}}^i[t]+1}}, ~~~ \forall ~ k \in \calI^i_{j^{\prime}}[t],\\
&{\bf M}_{n_{j^{\prime}j}, n_{kj^{\prime}}}[t] \triangleq \frac{\sfB^i_{(k,j^{\prime})}[t]}{d_{j^{\prime}}^{i}[t]+1},~~ \forall ~ k\in \calI^i_{j^{\prime}}[t], \\
& {\bf M}_{n_{j^{\prime}j}n_{j^{\prime}j}}[t] \triangleq 1- \sfB^i_{(j^{\prime}, j)}[t].
\end{align*}
and any other entry in ${\bf M}[t]$ be zero.
It is easy to check that the obtained matrix ${\bf M}[t]$ is column stochastic, and that 
\begin{align*}
    \bm{z}[t] = \pth{\bm{M}[t] \otimes \bm{I}}\bm{z}[t-1], ~~~ \forall ~ t \bmod \Gamma \not=0, 
\end{align*}
where $\bm{z}[t]\in \reals^{\tilde{N}d}$ is the vector that stacks all the local $z$'s. 
The update of the weight vector has the same matrix form since the update of value $z$ and weight $m$ are identical.

\paragraph{Global fusion iterations}
Fix $t$ be arbitrary iteration such that $t\mod \Gamma =0$. We construct matrix $\bm{M}$ in two steps. We let $\bar{\bm{M}}$ denote the matrix constructed the same way as above. Let $\bm{F}$ be the matrix that captures the mass push among the designated agents under the coordination of the parameter server. Specifically, 
\begin{align*}
\bm{F}_{j_0,j_0} &= \frac{M+1}{2M} ~~~~ \text{for each designated agent } j_0; \\
\bm{F}_{j_0,j^{\prime}_0} & = \frac{1}{2M} ~~~~ \text{for distinct designated agents } j_0, j^{\prime}_0,
\end{align*}
with all the other entries being zeros. Henceforth, we refer to matrix $\bm{F}$ as hierarchical fusion matrix. 
Clearly, $\bm{F}$ is a doubly-stochastic matrix. Hence, we define $\bm{M}$ as 
\begin{align}
\label{eq: fusion iteration}
\bm{M}[t] =\bm{F} \bar{\bm{M}}[t]. 
\end{align}

It is easy to see that the dynamics of $\bm{z}[t]$ for $t\mod \Gamma =0$ also obey 
\begin{align*}
    \bm{z}[t] = \pth{\bm{M}[t] \otimes \bm{I}}\bm{z}[t-1]. 
\end{align*}

\paragraph{Overall}
We have 
\begin{align*}
\bm{z}[t] & = \pth{\bm{M}[t] \otimes \bm{I}}\pth{\bm{M}[t-1] \otimes \bm{I}} \bm{z}[t-2] \\
& = \pth{\bm{M}[t] \otimes \bm{I}}\pth{\bm{M}[t-1] \otimes \bm{I}} \cdots \pth{\bm{M}[1] \otimes \bm{I}} \bm{z}[0]. 
\end{align*} 
That is, the evolution of $z$ is controlled by the matrix product $\bm{M}[t]\bm{M}[t-1]\cdots \bm{M}[1]$. 
In general, let ${\bf \Psi}(r,t)$ be the product of $t-r+1$ matrices
\begin{align*}
{\bf \Psi}(r,t)&\triangleq \prod_{\tau=r}^t\, {\bf M}^{\top}[\tau]={\bf M}^{\top}[r] {\bf M}^{\top}[r+1]\cdots {\bf M}^{\top}[t],
\end{align*}
where $r\le t$ with ${\bf \Psi}(t+1,t)\triangleq {\bf I}$ by convention, i.e., ${\bf \Psi}(r,t) = \pth{\bm{M}[t]\bm{M}[t-1]\cdots \bm{M}[1]}^{\top}$. 
Notably, ${\bf M}^{\top}[\tau]$ is row-stochastic for each $\tau$ of interest. 
Without loss of generality, let us fix an arbitrary bijection   
between $\{N+1, \cdots, \tilde{N}\}$ and $(j,j^{\prime})\in \calE_i$ for $i=1, \cdots, M$. For $j\in \calV_i$, we have 
\begin{align}
z_j^{i}[t] & = \sum_{j^{\prime}=1}^{\tilde{N}} z_{j^{\prime}}[0] {\bf \Psi}_{j^{\prime}j}(1,t) = \sum_{j^{\prime}=1}^{\tilde{N}} w_{j^{\prime}}^i {\bf \Psi}_{j^{\prime}j}(1,t),
\label{evoz}
\end{align}
where the last equality holds due to $z_{j}[0] =w_j^i$ for $j=1, \cdots, N$ and $z_{j}[0]=0\in \reals^d$ when $j>N$, i.e., when $j$ corresponds to an edge. 

\subsection{Auxiliary Lemmas}
\label{app: consensus: auxiliary lemma}
To show the convergence of Algorithm \ref{alg:push-sum hierarchical FL}, we investigate the convergence behavior of ${\bf \Psi}(r,t)$ (where $r\le t$) using ergodic coefficients and some celebrated results obtained by Hajnal \cite{Hajnal58}.
The remaining proof follows the same line as that in \cite{RobustAverageConsensus,vaidya2012robust}, and is presented below for completeness. 

Given a row stochastic matrix ${\bf A}$, coefficients of ergodicity $\delta({\bf A})$ is defined as:
\begin{align}
\delta({\bf A}) & \triangleq   \max_j ~ \max_{i_1,i_2}~\left | {\bf A}_{i_1 j}-{\bf A}_{i_2 j}\right |, \label{e_delta} \\
\lambda({\bf A}) & \triangleq   1 - \min_{i_1,i_2} \sum_j \min\{{\bf A}_{i_1 j}, {\bf A}_{i_2 j}\}. \label{e_lambda}
\end{align}
\begin{proposition}\cite{Hajnal58}
\label{prop: claim_delta}
For any $p$ square row stochastic matrices ${\bf Q}[1],{\bf Q}[2],\dots {\bf Q}[p]$, it holds that
\begin{align}
\delta({\bf Q}[1]{\bf Q}[2]\ldots {\bf Q}[p]) ~\leq ~
 \Pi_{k=1}^p ~ \lambda({\bf Q}[k]).
\end{align}
\end{proposition}

\vskip 0.5\baselineskip 
Proposition \ref{prop: claim_delta} implies that if $\lambda({\bf Q}[k])\leq 1-c$ for some $c>0$ and for all $1\le k\le p$, then $\delta({\bf Q}[1],{\bf Q}[2]\cdots {\bf Q}[p])$ goes to zero exponentially fast as $p$ increases.
The following lemmas are useful in the analysis of our follow-up algorithms. 

\begin{lemma}
\label{lm: matrix block rate}
For $r\le t$ such that $\lfloor t/2\Gamma \rfloor - \lceil r/2\Gamma \rceil\ge 0$, it holds that $\delta\pth{{\bf \Psi}(r, t)}\le \gamma^{(t-r)/2\Gamma},$
where $\gamma = 1-\frac{1}{4M^2} \pth{\min_{i\in [M]}\beta_i}^{2D^*B}$ as per Theorem \ref{thm: convergence of hierarchical FL}.
\end{lemma}
\begin{proof}
The following rewriting holds: 
\begin{equation*}
{\bf \Psi}(r, t) = {\bf \Psi}(r, \Gamma \lceil r/\Gamma\rceil)\pth{ \prod_{k=\lceil r/\Gamma \rceil}^{\lfloor t/\Gamma \rfloor-1} {\bf \Psi}(k\Gamma+1, (k+1)\Gamma)} \times {\bf \Psi}(\Gamma\lfloor t/\Gamma \rfloor +1, \, t). 
\end{equation*}
By Proposition \ref{prop: claim_delta}, we have 
\begin{equation*}
\delta\pth{{\bf \Psi}(r, t)} \le \prod_{k=\lceil r/2\Gamma \rceil}^{\lfloor t/2\Gamma \rfloor-1} \lambda\pth{{\bf \Psi}(2k\Gamma+1, 2(k+1)\Gamma)} \le \gamma^{(t-r)/2\Gamma}, 
\end{equation*}
where $\gamma := 1-\frac{1}{4M^2} \pth{\min_{i\in [M]}\beta_i}^{2D^*B}$.     
\end{proof}

\begin{lemma}
\label{lm: entire matrix lower bound}
Let $D^* := \max_{i\in [M]}D_i$. 
Choose $\Gamma = BD^*$. 
Suppose that $t-r+1\ge 2\Gamma$. Then every entry of the matrix product ${\bf \Psi}(r, t)$ is lower bounded by $\frac{1}{4M^2}\pth{\min_{i\in [M]}\beta_i}^{2D^*B}$. 
\end{lemma}
\begin{proof}
Observe that 
\begin{align*}
{\bf \Psi}(r,t) & = \bm{M}^{\top}[r]\cdots \bm{M}^{\top}[t-D_iB] \cdots \bm{M}^{\top}[t].    
\end{align*}
By Assumptions \ref{ass: link reliability} and \ref{ass: connectivity}, each subnetwork $S^i$ is strongly connected and each link is reliable at least once during $B$ consecutive iterations, it is true that every entry in the $i$-th block of matrix product $\bm{M}^{\top}[t-D_iB+1]\cdots \bm{M}^{\top}[t]$ is lower bounded by $\beta_i^{D_iB}$. We know that each of the remained matrices in ${\bf \Psi}(r,t)$ is row-stochastic. Hence, every entry in the block $i$ of ${\bf \Psi}(r,t)$ is lower bounded by $\beta_i^{D_iB}$.

By the construction of the fusion matrix $\bm{F}$ and the existence of self-loops, we know that during consecutive $2\Gamma$ iterations, from any node $j$, we can reach any other node in the hierarchical FL system. Let $j$ and $j^{\prime}$ be two arbitrary nodes possibly in different subnetworks. Let $S_i$ and $S_{i^{\prime}}$ be the subnetworks that $j$ and $j^{\prime}$ are in. 
It holds that
\begin{equation*}
{\bf \Psi}_{j,j^{\prime}}(t-2\Gamma +1,t) \ge \frac{1}{2M} \beta_{i^{\prime}}^{D_{i^{\prime}}B} \frac{1}{2M} \beta_i^{2D_iB} \ge \frac{1}{4M^2} (\min_{i\in [M]}\beta_i)^2D^*B, 
\end{equation*}
proving the lemma. 
\end{proof}

\subsection{Proof of Theorem \ref{rps convergence rate}}
\label{app:subsec:proof of theorem: consensus}
Notably, the update of the mass vector is $\bm{m}[t] = \pth{\bm{M}[t]\cdots \bm{M}[1]} \bm{m}[0] = \bm{\Psi}(1,t)\bm{m}[0],$
where 
$m_j[0]=1$ if $j\le N$ and $m_j[0]=0$ otherwise. 
\begin{align*}
\norm{\frac{z_j^i[t]}{m_j^i[t]} -\frac{1}{N}\sum_{i=1}^M\sum_{j=1}^{n_i} w_{j}^i}
& = \left \| \frac{N\sum_{i=1}^M\sum_{j^{\prime}=1}^{n_i}w_{j^{\prime}}^i{\bf \Psi}_{j^{\prime},j}(1,t)}{N\sum_{i=1}^M\sum_{j^{\prime}=1}^{n_i}{\bf \Psi}_{j^{\prime},j}(1,t)} - \frac{\sum_{i=1}^M\sum_{j^{\prime}=1}^{n_i} w_{j^{\prime}}^i \sum_{i=1}^M\sum_{k=1}^{n_i}{\bf \Psi}_{k,j}(1,t)}{N\sum_{i=1}^M\sum_{j^{\prime}=1}^{n_i}{\bf \Psi}_{j^{\prime},j}(1,t)} \right\|\\
& =\left \| \frac{\sum_{i=1}^M\sum_{j^{\prime}=1}^{n_i}w_{j^{\prime}}^i\sum_{i=1}^M\sum_{k=1}^{n_i}{\bf \Psi}_{j^{\prime},j}(1,t)}{N\sum_{i=1}^M\sum_{j^{\prime}=1}^{n_i}{\bf \Psi}_{j^{\prime},j}(1,t)}\right. \\
& \qquad \qquad \qquad -  \left. \frac{\sum_{i=1}^M\sum_{j^{\prime}=1}^{n_i} w_{j^{\prime}}^i \sum_{i=1}^M\sum_{k=1}^{n_i}{\bf \Psi}_{k,j}(1,t)}{N\sum_{i=1}^M\sum_{j^{\prime}=1}^{n_i}{\bf \Psi}_{j^{\prime},j}(1,t)} \right\|\\ 
& = \frac{\norm{\sum_{i=1}^M\sum_{j^{\prime}=1}^{n_i}w_{j^{\prime}}^i\sum_{i=1}^M\sum_{k=1}^{n_i}\pth{{\bf \Psi}_{j^{\prime},j}(1,t) -{\bf \Psi}_{k,j}(1,t)}}}{N\sum_{i=1}^M\sum_{j^{\prime}=1}^{n_i}{\bf \Psi}_{j^{\prime},j}(1,t)} \\
& \le \frac{\sum_{i=1}^M\sum_{j^{\prime}=1}^{n_i}\norm{w_{j^{\prime}}^i}\sum_{i=1}^M\sum_{k=1}^{n_i}\abth{{\bf \Psi}_{j^{\prime},j}(1,t) -{\bf \Psi}_{k,j}(1,t)}}{N\sum_{i=1}^M\sum_{j^{\prime}=1}^{n_i}{\bf \Psi}_{j^{\prime},j}(1,t)}\\
& \le \frac{\sum_{i=1}^M\sum_{j^{\prime}=1}^{n_i}\norm{w_{j^{\prime}}^i}\delta\pth{{\bf \Psi}(1, t)}}{\sum_{i=1}^M\sum_{j^{\prime}=1}^{n_i}{\bf \Psi}_{j^{\prime},j}(1,t)}. 
\end{align*}
By Lemmas \ref{lm: matrix block rate} and  \ref{lm: entire matrix lower bound}, we conclude that  
\begin{equation*}
\frac{\sum_{i=1}^M\sum_{j^{\prime}=1}^{n_i}\norm{w_{j^{\prime}}^i}\delta\pth{{\bf \Psi}(1, t)}}{\sum_{i=1}^M\sum_{j^{\prime}=1}^{n_i}{\bf \Psi}_{j^{\prime},j}(1,t)} \le \frac{4M^2\sum_{i=1}^M\sum_{j^{\prime}=1}^{n_i}\norm{w_{j^{\prime}}^i}}{\pth{\min_{i\in [M]}\beta_i}^{2D^*B}N}  \min\{1, \gamma^{\lfloor t/2\Gamma \rfloor - 1}\}. 
\end{equation*}

\section{State Estimation}
\label{app: state estimation}

\subsection{Proof of Proposition \ref{prop: lipschitz}}
\label{subsec: proof of lipschitz state}
By definition, $f(w) = \frac{1}{2}(w-w^*)^{\top}K(w-w^*) + \sum_{i=1}^M\sum_{j=1}^{n_i}\sigma_j^i$. 
We have 
\begin{align*}
\norm{f(w) - f(w^{\prime})}
& \le  \frac{1}{2}\norm{\pth{w-w^{\prime}}^{\top}K\pth{w- w^*}} + \frac{1}{2}\norm{\pth{w^{\prime}- w^*}^{\top}K\pth{w-w^{\prime}}} \\
& \le R_0\norm{K}\norm{w-w^{\prime}}. 
\end{align*}
Since $\pth{H^i_j}^{\top}H_j^i\succeq 0$, it holds that $\norm{H_j^{\top}H_j}\le \norm{K}.$ 
With similar argument, we can conclude that $f_j^i$ is also $L$-Lipschitz continuous with $L:=R_0\norm{K}$.

\subsection{Proof of Theorem \ref{thm: convergence of hierarchical FL}} 
Let $\bar{z}[t] = \frac{1}{N}\sum_{i=1}^M \sum_{j=1}^{n_i}z_j^i[t]$. Expanding $\bar{z}[t]$, we have
\begin{align*}
\bar{z}[t] & =  \frac{1}{N}\sum_{i=1}^M \sum_{j=1}^{n_i}\sum_{r=0}^{t-1} g_j^i[r]  \\
&= \frac{1}{N}\sum_{i=1}^M \sum_{j=1}^{n_i}\sum_{r=0}^{t-1} (H_j^i)^{\top}H_j^i \pth{w_j^i[r]-w^*} + \frac{1}{N}\sum_{i=1}^M \sum_{j=1}^{n_i} (H_j^i)^{\top} \sum_{r=0}^{t-1} \xi_j^i[r]. 
\end{align*}
It is worth noting that $\{w_j^i[t]\}_{t=0}^{\infty}$ is obtained under the stochastic gradient. 
Let $\{v[t]\}_{t=0}^{\infty}$ 
be the auxiliary sequence such that 
$v[t] : = \prod_{\calW}^{\varphi}\pth{\bar{z}[t], \eta[t-1]}$. Let $\hat{v}[t]:=\frac{1}{t}\sum_{r=0}^tv[r]$. 

Since $K$ is invertible (by Assumption \ref{ass: global observability}), the global objective $f$ has a unique minimizer $w^{*}$.
In addition, we have 
\begin{align*}
f\pth{\hat{w}_j^{i}[t]} - f(w^{*}) \ge\lambda_{\min}\pth{K}\norm{w_j^{i}[t]-w^{*}}^2.
\end{align*}
Thus, it remains to bound $f\pth{\hat{w}_j^{i}[t]} - f(w^{*})$. 
Fix a time horizon $t$. Since $w^{*}\in \calW$, we have 
\begin{align*}
f\pth{\hat{w}_j^{i}[t]} - f(w^{*})
& \le f\pth{\hat{v}[t]} - f(w^{*}) + L\norm{\hat{w}_j^{i}[t] - \hat{v}[t]}, ~~~ \text{by Proposition \ref{prop: lipschitz}} \\
& \overset{(a)}{\le} \frac{1}{t}\sum_{r=1}^t f(v[r]) - f(w^{*}) + \frac{1}{t}\sum_{r=1}^t L \norm{w_j^{i}[r] - v[r]} \\
& \overset{(b)}{\le}  \frac{1}{t}\sum_{r=1}^t f(v[r]) - f(w^{*}) + \frac{1}{t}\sum_{r=1}^t L \eta[r-1]\norm{\bar{z}[r] -\frac{z_j^{i}[r]}{m_j^{i}[r]}}, 
\end{align*}
where inequality (a) holds because of the convexity of $f$ and $\norm{\cdot}$, and inequality (b) follows from \cite[Lemma 2]{duchi2011dual}. 
For the first term, via similar argument, we have 
\begin{align*}
\frac{1}{t}\sum_{r=1}^t f(v[r]) - f(w^{*})  &=  \frac{1}{t}\sum_{r=1}^t \sum_{i=1}^M\sum_{j=1}^{n_i} f_j^i(w_i[r]) - f(w^{*}) + \frac{1}{t}\sum_{r=1}^t f(v[r]) - \frac{1}{t}\sum_{r=1}^t \sum_{i=1}^M\sum_{j=1}^{n_i} f_j^i(w_i[r]) \\
&\le \frac{1}{t}\sum_{r=1}^t \sum_{i=1}^M\sum_{j=1}^{n_i} f_j^i(w_j^i[r]) - \sum_{i=1}^M\sum_{j=1}^{n_i} f_j^i(w^{*}) + \frac{1}{t}\sum_{r=1}^t \sum_{i=1}^M\sum_{j=1}^{n_i} L\eta[r-1]\norm{\bar{z}[r] -\frac{z_i[r]}{m_i[r]}}. 
\end{align*}

For ease of notation, let $g_j^{i\prime} = \nabla f_j^i(w_j^i[r])$. Since $f_j^i$ is convex, we have  
\begin{align*}
f_j^i(w_j^i[r]) - f_j^i(w^{*}) &\le \iprod{g_j^{i\prime}[r]}{w_j^i[r] - w^*} \\
& = \iprod{g_j^i[r]}{w_j^i[r] - w^*} + \iprod{g_j^{i\prime}[r] - g_j^i[r]}{w_j^i[r] - w^*}.  
\end{align*}
We have,   
\begin{align}
\label{eq: state: 111}
\nonumber
&\frac{1}{t}\sum_{r=1}^t \sum_{i=1}^M\sum_{j=1}^{n_i} f_j^i(w_j^i[r]) - \sum_{i=1}^M\sum_{j=1}^{n_i} f_j^i(w^*) \\
\nonumber
& \le  \frac{1}{t}\sum_{r=1}^t \sum_{i=1}^M\sum_{j=1}^{n_i} \iprod{g_j^i[r]}{w_j^i[r] - w^*} + \frac{1}{t}\sum_{r=1}^t \sum_{i=1}^M\sum_{j=1}^{n_i}\iprod{g_j^{i\prime}[r] - g_j^i[r]}{w_j^i[r] - w^*} \\
& \le \frac{1}{2t}\sum_{i=1}^M\sum_{j=1}^{n_i}\sum_{r=1}^t \eta[r-1] L_0^2  + \frac{N}{t\eta[t]}\varphi(w^*) + \frac{1}{t}\sum_{i=1}^M\sum_{j=1}^{n_i} \sum_{r=1}^t \iprod{g_j^{i\prime}[r] - g_j^i[r]}{w_j^i[r] - w^*}, 
\end{align}
where the last inequality holds from Proposition \ref{prop: bounded stochastic gradient} and \cite[Lemma 3]{duchi2011dual}. 
In addition, it can be easily checked by definition that $\iprod{g_j^{i\prime}[r] - g_j^i[r]}{w_j^i[r] - w^*}$ is a martingale. 
By the Cauchy–Schwarz inequality, we have  
\begin{align*}
\abth{\iprod{g_j^{i\prime}[r] - g_j^i[r]}{w_j^i[r] - w^*}} \le 2LR_0.     
\end{align*}
That is, $\iprod{g_j^{i\prime}[r] - g_j^i[r]}{w_j^i[r] - w^*}$ is a bounded difference martingale. By Azuma's inequality, with probability at least $1-\delta$, it holds that 
\begin{align*}
\frac{1}{Nt}\sum_{r=1}^t \sum_{i=1}^M\sum_{j=1}^{n_i} \iprod{g_j^{i\prime}[r] - g_j^i[r]}{w_j^i[r] - w^*} \le 4LR_0\sqrt{\frac{\log \frac{1}{\delta}}{t}}.     
\end{align*}
Hence, with probability at least $(1-\delta)$, Eq.\eqref{eq: state: 111} is upper bounded as 
\[
 \frac{1}{2t}\sum_{i=1}^M\sum_{j=1}^{n_i}\sum_{r=1}^t \eta[r-1] L_0^2  + \frac{N}{t\eta[t]}\varphi(w^*) + 4NLR_0\sqrt{\frac{\log 1/\delta}{t}}. 
\]
Using steps similar to the proof in Theorem \ref{rps convergence rate}, we are able to show  
\begin{align*}
\norm{\bar{z}[r] -\frac{z_j^{i}[r]}{m_j^{i}[r]}}\le \frac{4M^2L\gamma^{1/2\Gamma}}{N\pth{1-\gamma^{1/2\Gamma}}\pth{\min_{i\in [M]}\beta_i}^{2D^*B}}, 
\end{align*}
proving the theorem.

\section{Tracking Analysis} 
\label{app: tracking analysis}

\subsection{Dynamic Matrix Representation} \label{tracking_prelim}
Recall that $\tilde{N}$ is the number of nodes in the $M$ augmented graphs, one for each network.  
Let $\bm{g}[t]\in \reals^{d\tilde{N}}$ be the vector that stacks the local stochastic gradients computed by the $N$ agents with $g_j[t] = \bm{0}$ if $j$ corresponds to an edge.  

\noindent\underline{Fix $t$ be arbitrary iteration such that $t\mod \Gamma \not=0$.} 
Fix a network $S_i$. For each $j^{\prime} \in \calI_j[t]$, it holds that 
\begin{equation} 
\label{eq: tracking: receive records}
\rho^+_{j^{\prime}j}[t] - \rho_{j^{\prime} j}[t-1] \\
= B_{j^{\prime} j}[t]z_{n_{j^{\prime}j}}[t-1] + B_{j^{\prime} j}[t] \frac{z_{j^{\prime}}[t-1]}{d_{j^{\prime}}[t]+1}. 
\end{equation}

For each $j\in \cup_{i=1}^M \calV_i$, we have 
\begin{multline*}
z_j[t] =
A\frac{z_j^+[t]}{d_j[t]+1} - g_j[t] = A \frac{1}{d_j[t]+1}\times \pth{\frac{z_j[t-1]}{d_j[t]+1} + \sum_{j^{\prime} \in \calI_j[t]} (\rho^+_{j^{\prime}j}[t] - \rho_{j^{\prime} j}[t-1])}  - g_j[t] \\
= \frac{Az_j[t-1]}{\pth{d_j[t]+1}^2} + \sum_{j^{\prime} \in \calI_j[t]}\frac{B_{j^{\prime} j}[t]}{d_j[t]+1} Az_{n_{j^{\prime}j}}[t-1]
+ \sum_{j^{\prime} \in \calI_j[t]}\frac{B_{j^{\prime} j}[t]}{\pth{d_j[t]+1}\pth{d_{j^{\prime}}[t]+1}}Az_{j^{\prime}}[t-1] - g_j[t]. 
\end{multline*}

For each edge $(j^{\prime}, j)$, we have 
\begin{align}
\label{eq: tracking: update of virtual nodes}
\nonumber z_{n_{j^{\prime}j}[t]} &= \sigma_{j^{\prime}}[t] - \rho_{j^{\prime}j}[t] = A\pth{\sigma_{j^{\prime}}^+[t] + \frac{z_{j^{\prime}}^+[t]}{d_{j^{\prime}}[t] +1}} - A\rho_{j^{\prime}j}^+[t] \\
\nonumber
&= A\qth{\pth{1 -  B_{j^{\prime} j}[t]} \pth{z_{n_{j^{\prime} j}}[t-1] + \frac{z_{j^{\prime}}[t-1]}{d_{j^{\prime}}[t]+1}} + \frac{z_{j^{\prime}}^+[t]}{d_{j^{\prime}}[t]+1} } \\
\nonumber
&= \pth{1 -  B_{j^{\prime} j}[t]} Az_{n_{j^{\prime} j}}[t-1] + \pth{\frac{1 -  B_{j^{\prime} j}[t]}{d_{j^{\prime}}[t]+1} + \frac{1}{\pth{d_{j^{\prime}}[t]+1}^2}}  Az_{j^{\prime}}[t-1]  \\
&\qquad + \sum_{k \in \calI_{j^{\prime}}} \frac{B_{kj^{\prime}}[t]}{d_{j^{\prime}}[t]+1}Az_{n_{kj^{\prime}}}[t-1] + \sum_{k \in \calI_{j^{\prime}}} \frac{B_{kj^{\prime}}[t]}{\pth{d_{j^{\prime}}[t]+1} \pth{d_{k}[t]+1}}Az_k[t-1].    
\end{align}
Hence, we have 
\begin{equation*}
\bm{z}[t] = \pth{\bm{M}[t] \otimes \bm{I}}\tilde{\bm{z}}[t-1] - \bm{g}[t],     
\end{equation*}
where 
\begin{equation*}
    \qth{\tilde{\bm{z}}[t-1]}^{\top} = \pth{z_1[t-1]^{\top} A^{\top},  \cdots,   z_{\tilde{N}}[t-1]^{\top} A^{\top}}
\end{equation*}
i.e., $\tilde{\bm{z}}[t-1] = \pth{\bm{I}\otimes A}\bm{z}[t-1]$.
Similarly, we can show the same matrix representation holds for any \underline{$t$ be arbitrary iteration such that $t\mod \Gamma =0$.}

Following from the fact that $\pth{\bm{A}\otimes \bm{B}}\pth{\bm{C}\otimes\bm{D}} = \pth{\bm{A}
\bm{C}}\otimes \pth{\bm{B}\bm{D}}$ with matrices $\bm{A}, \bm{B}, \bm{C},$ and $\bm{D}$ of proper dimensions so that the relevant matrix product is well-defined, we have 
\begin{align}
\label{eq: tracking matrix dynammics}
\nonumber 
\bm{z}[t]  &= \pth{\bm{M}[t] \otimes \bm{I}}\pth{\bm{I}\otimes A}\bm{z}[t-1] - \bm{g}[t] \\
 &= \pth{\bm{M}[t] \otimes \bm{A}}\bm{z}[t-1] - \bm{g}[t].  
\end{align} 
Unrolling Eq.\eqref{eq: tracking matrix dynammics}, we get 
\begin{align}
\nonumber
\bm{z}[t] &= \pth{\bm{M}[t] \otimes A}\bm{z}[t-1] - \bm{g}[t] \\
\nonumber
&\overset{(a)}{=}  \pth{\bm{M}[t] \otimes A} \cdots  \pth{\bm{M}[1] \otimes A}\bm{z}[0]- \sum_{r=1}^t \pth{\pth{\bm{M}[t] \otimes A} \cdots  \pth{\bm{M}[r+1] \otimes A}}\bm{g}[r]\\
&= -\sum_{r=1}^t \pth{\pth{\bm{M}[t] \otimes A} \cdots  \pth{\bm{M}[r+1] \otimes A}}\bm{g}[r],     
\end{align}
where in equality (a) we use the convention that $\pth{\bm{M}[t] \otimes A} \cdots  \pth{\bm{M}[r+1] \otimes A} = \bm{I}_{d\tilde{N}}$. 
Repeatedly applying the fact that $\pth{\bm{A}\otimes \bm{B}}\pth{\bm{C}\otimes\bm{D}} = \pth{\bm{A}
\bm{C}}\otimes \pth{\bm{B}\bm{D}}$, we obtain 
\begin{equation*}
\bm{z}[t] = -\sum_{r=1}^t \pth{\pth{\bm{M}[t]\cdots \bm{M}[r+1]}\otimes A^{t-r}}\bm{g}[r]. 
\end{equation*}
Notably, the update of the mass vector remains the same as that for Algorithm \ref{alg:push-sum hierarchical FL}, i.e., 
\begin{align*}
 \bm{m}[t] = \pth{\bm{M}[t]\cdots \bm{M}[1]} \bm{m}[0], 
\end{align*}
where 
$m_j[0]=1$ if $j\le N$ and $m_j[0]=0$ otherwise. 

\subsection{Proof of Theorem \ref{thm: convergence of hierarchical FL: tracking}} 
\label{tracking_proof analysis}
Recall that we index the nodes in the augmented graphs from $1$ to $\tilde{N}$ with nodes $1$ to $N$ corresponding to the actual nodes (i.e., $j\in \cup_{i=1}^M\calV_i$) and nodes $N$ to $\tilde{N}$ corresponding to the virtual nodes (i.e., $j\in \cup_{i=1}^M \calE_i$).  For each agent $j\in \cup_{i=1}^M\calV_i$,  we have  
\begin{align*}
z_j[t] = -\sum_{r=1}^{t} \sum_{j^{\prime}=1}^{\tilde{N}} A^{t-r}g_{j^{\prime}}[r]\bm{\Psi}_{j^{\prime}j}(r+1,t),  
\end{align*}
where $g_j[t]$ is the stochastic gradient of Eq.\,\eqref{eqn: local asymptotic function} 
which can be rewritten as 
\begin{align*}
g_j[t] & = H_j^\intercal \pth{H_j  w_j[t-1] -H_jw^*[t] - \xi_j[t-1]} \\ 
& = H_j^\intercal H_j\pth{ w_j[t-1] - w^*[t-1]}  - H_j^\intercal \xi_j[t-1]. 
\end{align*}
In addition,
\[
m_j[t] = \sum_{j^{\prime}=1}^{\tilde{N}} \bm{\Psi}_{j^{\prime}j}(1,t) m_j[0] = \sum_{j^{\prime}=1}^{N} \bm{\Psi}_{j^{\prime}j}(1,t).
\]
The evolution of $\bar{z}$  can be formally described as 
\begin{align*}
\bar{z}[t+1] &= \frac{1}{N}\sum_{j=1}^{\tilde{N}}z_j[t+1]  \\
&= -\frac{1}{N}\sum_{r=1}^{t+1}  A^{t+1-r} \eta[r]\sum_{j^{\prime}=1}^{N}g_{j^{\prime}}[r] \\
&= -\frac{1}{N} \pth{A \sum_{r=1}^{t}  A^{t-r} \eta[r]\sum_{j^{\prime}=1}^{N}g_{j^{\prime}}[r] + \eta[t+1]\sum_{j^{\prime}=1}^{N}g_{j^{\prime}}[t+1]} \\
&= A \bar{z}[t] - \eta[t+1]\frac{1}{N}\sum_{j^{\prime}=1}^{N}g_{j^{\prime}}[t+1]. 
\end{align*}
Let $\tilde{w}_j[t] : = \frac{z_j[t]}{m_j[t]}.$ 
By non-expansion property of projection onto a convex and compact set, we have 
\[
\norm{w_j[t]-w^*[t]} \le \norm{\Tilde{w}_j[t]-w^*[t]}. 
\]
Note that 
\begin{equation}
\label{eq: tracking average cumultive error}
\tilde{w}_j[t] - w^*[t]
=  \underbrace{\pth{\bar{z}[t] - w^*[t]}}_{(a)} + \underbrace{\pth{\tilde{w}_j[t] - \bar{z}[t]}}_{(b)}.
\end{equation}

\noindent{\bf Bounding (b):}
\begin{align*}
\norm{\tilde{w}_j[t] - \bar{z}[t]} &=  \norm{\frac{z_j[t]}{m_j[t]} - \bar{z}[t]} \\
&= \left \| \frac{\sum_{r=1}^{t-1}\sum_{j^{\prime}=1}^{N} A^{t-r}\eta[r]g_{j^{\prime}}[r] {\bf \Psi}_{j^{\prime},j}(r,t)}{\sum_{j^{\prime}=1}^{N}{\bf \Psi}_{j^{\prime},j}(1,t)} -\frac{1}{N}\sum_{r=1}^{t}  A^{t-r}\eta[r]\sum_{j^{\prime}=1}^{N}g_{j^{\prime}}[r] \right \|_2 \\
&= \left \| \frac{\sum_{k=1}^N\sum_{r=1}^{t-1}\sum_{j^{\prime}=1}^{N} A^{t-r}\eta[r]g_{j^{\prime}}[r] {\bf \Psi}_{j^{\prime},j}(r,t)}{N\sum_{j^{\prime}=1}^{N}{\bf \Psi}_{j^{\prime},j}(1,t)}  \right. \\
&\qquad \qquad \quad \left. -\frac{\sum_{r=1}^{t}  A^{t-r}\eta[r]\sum_{j^{\prime}=1}^{N}g_{j^{\prime}}[r]\sum_{k=1}^{N}{\bf \Psi}_{k,j}(1,t)}{N\sum_{j^{\prime}=1}^{N}{\bf \Psi}_{j^{\prime},j}(1,t)} \right \|_2 \\
&\overset{(a)}{\le} \frac{4M^2L_0\sum_{r=0}^{t-1} \sum_{j^{\prime}=1}^{N} \norm{A}^{t-r}\eta[r] \delta\pth{{\bf \Psi}(r,t)}}{ N  \pth{\min_{i\in [M]}\beta_i}^{2D^*B}} \\
&\overset{(b)}{\le} \frac{4M^2L_0\sum_{r=0}^{t-1}  \norm{A}^{t-r}\eta[r] \gamma^{\frac{t-r}{\Gamma}}}{\pth{\min_{i\in [M]}\beta_i}^{2D^*B}}\\
&= \frac{4M^2L_0\sum_{r=0}^{t-1} \eta[r] \pth{\norm{A}\gamma^{\frac{1}{\Gamma}}}^{t-r}}{\pth{\min_{i\in [M]}\beta_i}^{2D^*B}}
\end{align*}
where inequality (a) follows from Proposition \ref{prop: bounded stochastic gradient},  the definition of $\delta\pth{\cdot}$ as per Eq.\eqref{e_delta}, and Lemma \ref{lm: entire matrix lower bound}, and inequality (b) follows from Lemma \ref{lm: matrix block rate}. 
For ease of exposition, let $b= \norm{A}\gamma^{\frac{1}{\Gamma}}$. Let $r^*\in \{1, 2, \cdots, t-1\}$. 
It holds that 
\begin{equation*}
\sum_{r=0}^{t-1} \eta[r] \pth{\norm{A}\gamma^{\frac{1}{\Gamma}}}^{t-r}   
= \sum_{r=0}^{t-1} \eta[r] b^{t-r} 
= \frac{1}{\norm{K}}\pth{\underbrace{b^t + \sum_{r=1}^{r^*} \frac{1}{r} b^{t-r}}_{(i)} + \underbrace{\sum_{r=r^* + 1}^{t-1} \frac{1}{r} b^{t-r}}_{(ii)}}
\end{equation*}
Term $(i)$ can be upper bounded as 
\begin{equation*}
b^t + \sum_{r=1}^{r^*} \frac{1}{r} b^{t-r}  \le b^t + \sum_{r=1}^{r^*} b^{t-r} = \sum_{r=0}^{r^*} b^{t-r} \le \frac{b^{t-r^*}}{1-b}
\end{equation*}
and term $(ii)$ can be upper bounded as 
\begin{equation*}
\sum_{r=r^* + 1}^{t-1} \frac{1}{r} b^{t-r} 
\le \sum_{r=r^* + 1}^{t-1} \frac{1}{r^*+1} b^{t-r}
= \frac{1}{r^*+1} \sum_{r=r^* + 1}^{t-1} b^{t-r} 
\le \frac{1}{(r^*+1)(1-b)}
\end{equation*}
Choosing $r^* = 1/2 t$, when 
\begin{equation}
\label{eq: threshold time}
t \ge \frac{2}{\log 1/b} \log \pth{\frac{2}{\log 1/b}}  ~ \triangleq ~ t_0, 
\end{equation}
where the base of the log is $2$, both of the upper bounds above can be further upper bounded as $\frac{2}{(1-b)t}$. Thus, 
\begin{equation*}
\sum_{r=0}^{t-1} \eta[r] \pth{\norm{A}\gamma^{\frac{1}{\Gamma}}}^{t-r} \le \frac{4}{(1-b)t}~~~ \qquad \forall t\ge t_0. 
\end{equation*}
For $t<t_0$, it holds that 
\begin{equation*}
\sum_{r=0}^{t-1} \eta[r] \pth{\norm{A}\gamma^{\frac{1}{\Gamma}}}^{t-r} 
\le \sum_{r=0}^{t-1} \eta[r] 
\le \sum_{r=0}^{t-1}
\le \sum_{r=0}^{t_0-1}
=t_0. 
\end{equation*}
Therefore, 
\begin{equation*}
\norm{\tilde{w}_j[t] - \bar{z}[t]} 
\le 
\begin{cases}
\frac{16M^2L_0}{\pth{\min_{i\in [M]}\beta_i}^{2D^*B}\norm{K}(1-b)t} ~ & \text{if } ~ t \ge t_0; \\ 
\frac{4M^2L_0t_0}{\pth{\min_{i\in [M]}\beta_i}^{2D^*B}\norm{K}} ~ & \text{if } ~ t < t_0, 
\end{cases}
\end{equation*}
proving the first part of Theorem \ref{thm: convergence of hierarchical FL}. 

\noindent{\bf Bounding (a):} 
\begin{align}
\label{eq: tracking: 111}
\nonumber \bar{z}[t] - w^*[t] &= A\bar{z}[t-1] - \eta[t] \frac{1}{N}\sum_{j=1}^N \left(H_j^{\top}H_j\pth{w_j[t-1] -w^*[t-1]} + H_j^{\top}\xi_j[t-1] \vphantom{w_j[t-1]} \right) - w^*[t] \\
\nonumber
&=  A\pth{\bar{z}[t-1] - w^*[t-1]} - \eta[t] \frac{1}{N}\sum_{j=1}^N  H_j^{\top}\xi_j[t-1]\\
&\qquad  - \eta[t](\frac{1}{N}\sum_{j=1}^N H_j^{\top}H_j \pth{w_j[t-1] -w^*[t-1]}). 
\end{align}
Adding and subtracting $\bar{z}[t-1]$ in each of the summand in the last term of Eq.\eqref{eq: tracking: 111}, and regrouping the terms, we get  
\begin{align*}
\bar{z}[t] - w^*[t] 
&= \pth{A - \frac{\eta[t]}{N}K}\pth{\bar{z}[t-1] - w^*[t-1]} \\
& \qquad - \frac{\eta[t]}{N}K\pth{w_j[t-1] -\bar{z}[t-1]} \\
& \qquad - \frac{\eta[t]}{N}\sum_{j=1}^N  H_j^{\top}\xi_j[t-1]. 
\end{align*}
Let $C[t-1] = \eta[t] \frac{1}{N}\sum_{j=1}^N H_j^{\top} H_j\pth{w_j[t-1] -\bar{z}[t-1]}$
and $W[t-1] =  \eta[t] \frac{1}{N}\sum_{j=1}^N  H_j^{\top}\xi_j[t-1]$. 
Notably, $\expect{W[t-1]} = \expect{\eta[t] \frac{1}{N}\sum_{j=1}^N  H_j^{\top}\xi_j[t-1]} = \bm{0}$. We unroll 
$\bar{z}[t] - w^*[t] $ as
\begin{align*}
\bar{z}[t] - w^*[t] &= \pth{A - \frac{\eta[t]}{N}K}\pth{\bar{z}[t-1] - w^*[t-1]} - C[t-1] - W[t-1] \\
&= \underbrace{\pth{A - \frac{\eta[t]}{N}K} \cdots \pth{A - \frac{\eta[1]}{N}K}\pth{\bar{z}[0] - w^*[0]}}_{(A)} \\
&- \underbrace{\sum_{r=2}^{t+1} \pth{A - \frac{\eta[t]}{N}K} \cdots \pth{A - \frac{\eta[r]}{N}K} C[r-2] }_{(B)}\\
& - \underbrace{\sum_{r=2}^{t+1} \pth{A - \frac{\eta[t]}{N}K} \cdots \pth{A - \frac{\eta[r]}{N}K} W[r-2]}_{(C)}.       
\end{align*}

Since $A$ is symmetric, it holds that 
\[
A = U\Lambda U^{\top}, 
\]
where $U\in \reals^{d\times d}$ is the square $d\times d$ matrix whose $i$-th column is the $i$-th eigenvector of $A$, and $\Lambda$ is the diagonal matrix whose diagonal elements are the corresponding eigenvalues. 
Thus, 
\[
A - \eta[t]K = U\Lambda U^{\top} - \eta[t]K = U\pth{\Lambda - \eta[t] U^{\top}KU}U^{\top}. 
\]
For any $r$, it holds that 
\begin{align*}
&\pth{A - \frac{\eta[t]}{N}K} \pth{A - \frac{\eta[t-1]}{N}K}\cdots \pth{A - \frac{\eta[r]}{N}K}\\
&= U\pth{\Lambda - \eta[t] U^{\top}KU}U^{\top} U\pth{\Lambda - \eta[t-1] U^{\top}KU}U^{\top} \cdots U\pth{\Lambda - \eta[r] U^{\top}KU}U^{\top}\\
&= U\pth{\Lambda - \eta[t] U^{\top}KU}\pth{\Lambda - \eta[t-1] U^{\top}KU} \cdots  \pth{\Lambda - \eta[r] U^{\top}KU}U^{\top}. 
\end{align*}
Thus, 
\begin{equation*}
\norm{\pth{A - \frac{\eta[t]}{N}K} \pth{A - \frac{\eta[t-1]}{N}K}\cdots \pth{A - \frac{\eta[r]}{N}K}} \le  \prod_{\tau=r}^t \norm{\Lambda - \eta[\tau] U^{\top}KU}.   
\end{equation*}
 
Notably, $U$ is a rotation matrix. Thus, $U^{\top}KU$ and $K$ share the same set of eigenvalues. By definition, we have 
\begin{align*}
\norm{\Lambda - \eta[\tau] U^{\top}KU} &= \sup_{v\in \calS^d} v^{\top} \pth{\Lambda - \eta[\tau] U^{\top}KU} v\\
&= \sup_{v\in \calS^d} v^{\top} \Lambda v - \eta[\tau] \inf_{v\in \calS^d} v^{\top}U^{\top}KU v \\
&\le 1 - \eta[\tau] \lambda_d. 
\end{align*}

So for $t\ge r$ 
\begin{align*}
\norm{\pth{A - \frac{\eta[t]}{N}K} \cdots \pth{A - \frac{\eta[r]}{N}K}} 
&\le \prod_{\tau=r}^t \pth{1-\eta[\tau] \lambda_d} \\
&= \exp\pth{\sum_{\tau=r}^t \ln \pth{1-\eta[\tau] \lambda_d}} \\
&\le \exp\pth{\sum_{\tau=r}^t -\eta[\tau] \lambda_d} \\
&= \exp\pth{\frac{\lambda_1}{\lambda_d}} \exp\pth{-\sum_{\tau=r}^t \frac{1}{\tau}} \\
&\le \exp\pth{\frac{\lambda_1}{\lambda_d}} \exp\pth{ -\log (t) + \log (r-1)}\\
&= \exp\pth{\frac{\lambda_1}{\lambda_d}} \frac{r-1}{t}.    
\end{align*}

Besides, 
\begin{equation*}
C[t-1] = \eta[t] \frac{1}{N}\sum_{j=1}^N H_j^{\top} H_j\pth{w_j[t-1] -\bar{z}[t-1]}.  
\end{equation*}

\noindent{\bf Bounding (A)}
\begin{align*}
&\left\|  \pth{A - \frac{\eta[t]}{N}K} \cdots \pth{A - \frac{\eta[1]}{N}K}\pth{\bar{z}[0] - w^*[0]}  \right\|_2 \\
&\le \left \| \pth{A - \frac{\eta[t]}{N}K} \cdots  \pth{A - \frac{\eta[1]}{N}K}\right \|_2 \norm{\bar{z}[0] - w^*[0]}\\
&= \left \| \pth{A - \frac{\eta[t]}{N}K} \cdots \pth{A - \frac{\eta[1]}{N}K} \right \|_2 \norm{w^*[0]}\\
&\le  \norm{w^*[0]} \exp\pth{\frac{\lambda_1}{\lambda_d}}\frac{1}{t}.  
\end{align*}

\noindent{\bf Bounding (B)}
\begin{align*}
&\norm{\sum_{r=2}^{t+1} \pth{A - \frac{\eta[t]}{N}K} \cdots \pth{A - \frac{\eta[r]}{N}K} C[r-2]} \\
&\le \sum_{r=2}^{t+1} \norm{\pth{A - \frac{\eta[t]}{N}K} \cdots \pth{A - \frac{\eta[r]}{N}K}}\norm{C[r-2]} \\
&\le \exp\pth{\frac{\lambda_1}{\lambda_d}}\sum_{r=2}^{t+1} \frac{r-1}{t}  \frac{1}{\norm{K} (r-1)}\frac{1}{N}\sum_{j=1}^N \norm{H_j^{\top} H_j} \max_{j}\norm{w_j[r-2]-\bar{z}[r-2]}\\
&\le \frac{1}{t} \exp\pth{\frac{\lambda_1}{\lambda_d}}\sum_{r=2}^{t+1} \max_{j}\norm{w_j[r-2]-\bar{z}[r-2]}\\
& \le \frac{1}{t} \exp\pth{\frac{\lambda_1}{\lambda_d}}\sum_{r=2}^{t+1} \max_{j}\norm{\tilde{w}_j[r-2]-\bar{z}[r-2]}, 
\end{align*}
where the last inequality holds from the non-expansion property of projection. 
Suppose that $t\ge t_0$. We have 
\begin{align*}
&\frac{1}{t} \exp\pth{\frac{\lambda_1}{\lambda_d}}\sum_{r=2}^{t+1} \max_{j}\norm{\tilde{w}_j[r-2]-\bar{z}[r-2]}\\
& \le \frac{\exp\pth{\lambda_1/\lambda_d}4M^2Lt^2_0}{\pth{\min_{i\in [M]}\beta_i}^{2D^*B}\norm{K}}\frac{1}{t} + \frac{\exp\pth{\lambda_1/\lambda_d} 16M^2L}{\pth{\min_{i\in [M]}\beta_i}^{2D^*B}\norm{K}(1-b)}\frac{\log(t+1)}{t}.  
\end{align*}

\noindent{\bf Bounding (C), Handling noises.}
We will use McDiarmid's inequality to derive high probability bound on term (C).   
 
Let's perturb the observation noise of agents at time $r^{\prime}-2$. 
It is easy to see that the difference on each of the coordinates is upper bounded by $\exp\pth{\lambda_1/\lambda_d} \frac{2B_0}{\norm{K}} \frac{1}{t}. $  

By McDiarmid's inequality, we obtain that 
with probability at most $\delta/d$, 
\begin{equation*}
\left [ \sum_{r=2}^{t+1} \pth{A - \frac{\eta[t]}{N}K} \cdots \pth{A - \frac{\eta[r]}{N}K}W[r-2] \right ]^i \ge \exp\pth{\lambda_1/\lambda_d} \frac{2B_0}{\norm{K}} \sqrt{\frac{\log d/\delta}{2t}}.  
\end{equation*}
Therefore, we conclude that with probability at least $1-\delta$, 
\begin{equation*}
\norm{\sum_{r=2}^{t+1} \pth{A - \frac{\eta[t]}{N}K} \cdots \pth{A - \frac{\eta[r]}{N}K}W[r-2]} \le \sqrt{\frac{d}{2t}\log (d/\delta)}\exp\pth{\lambda_1/\lambda_d} \frac{2B_0}{\norm{K}}.  
\end{equation*}

Combining the bounds on terms  (A), (B), and (C), and (b), we conclude the theorem.